%% file: main.tex
\documentclass{article}

\PassOptionsToPackage{numbers, compress}{natbib}
\usepackage[preprint]{neurips_2025}

\usepackage[utf8]{inputenc} % allow utf-8 input
\usepackage[T1]{fontenc}    % use 8-bit T1 fonts
\usepackage[hidelinks]{hyperref}
\usepackage{url}            % simple URL typesetting
\usepackage{booktabs}       % professional-quality tables
\usepackage{amsfonts}       % blackboard math symbols
\usepackage{nicefrac}       % compact symbols for 1/2, etc.
\usepackage{microtype}      % microtypography
\usepackage[dvipsnames]{xcolor}

\hypersetup{
    colorlinks,
    linkcolor=RoyalBlue,%{red!50!black},
    citecolor=RoyalBlue, %{blue!50!black},
    urlcolor=RoyalBlue, %{blue!80!black}
}

\usepackage[textsize=small]{todonotes}
\usepackage{wrapfig}
\usepackage{graphicx}
\usepackage{amsmath}
\usepackage{amssymb}
\usepackage{adjustbox}
\usepackage{tikz}
\usepackage{tikz-3dplot}
\usepackage{pgfplots}
\usepackage{lipsum}  
\usepackage{amsthm}
\usepackage{mathtools}
\usepackage[linesnumbered,ruled,vlined,nosemicolon]{algorithm2e}
\usepackage{algorithmic}
\usepackage[T1]{fontenc}
\usepackage{tikz-cd}
\usepackage{capt-of}
\usepackage{float}
\usepackage{iac_pkg}
\usepackage[capitalise]{cleveref}
\usepackage{enumitem}
\usepackage{subcaption} % Aggiungi questo nel preambolo
\usepackage{multirow}
\usepackage{colortbl}

\usepackage{arydshln}

\newcommand{\dataset}[1]{#1} % TODO: define a common format for all datasets
\newcommand{\model}[1]{#1} % TODO: define a common format for all models

\makeatletter
\def\adl@drawiv#1#2#3{%
        \hskip.5\tabcolsep
        \xleaders#3{#2.5\@tempdimb #1{1}#2.5\@tempdimb}%
                #2\z@ plus1fil minus1fil\relax
        \hskip.5\tabcolsep}
\newcommand{\cdashlinelr}[1]{%
  \noalign{\vskip\aboverulesep
           \global\let\@dashdrawstore\adl@draw
           \global\let\adl@draw\adl@drawiv}
  \cdashline{#1}
  \noalign{\global\let\adl@draw\@dashdrawstore
           \vskip\belowrulesep}}
\makeatother

\input{don_macros.tex}

\usetikzlibrary{shapes.geometric, arrows.meta, positioning, decorations.pathreplacing, spy,calc, shadows, shadows.blur, angles, spath3}

\tikzstyle{data} = [rectangle, rounded corners, minimum width=2cm, minimum height=1cm, text centered, draw=black, fill=green!30]
\tikzstyle{model} = [rectangle, minimum width=2cm, minimum height=1cm, text centered, draw=black, fill=orange!30]
\tikzstyle{arrow} = [thick,->,>=stealth]

\tikzset{%
    add/.style args={#1 and #2}{
        to path={%
 ($(\tikztostart)!-#1!(\tikztotarget)$)--($(\tikztotarget)!-#2!(\tikztostart)$)%
  \tikztonodes},add/.default={.2 and .2}}
}

\tikzset{%
  mark coordinate/.style={inner sep=0pt,outer sep=0pt,minimum size=2pt,
    fill=black,circle}%
}

\tdplotsetmaincoords{60}{115}
\pgfplotsset{compat=newest}

\pgfmathdeclarefunction{sphereX}{2}{%
    \pgfmathparse{sin(#2)*cos(#1)}%
}
\pgfmathdeclarefunction{sphereY}{2}{%
    \pgfmathparse{sin(#2)*sin(#1)}%
}
\pgfmathdeclarefunction{sphereZ}{2}{%
    \pgfmathparse{cos(#2)}%
}

\pgfmathdeclarefunction{stereographicProjection}{2}{%
    \pgfmathparse{#1/(1-#2)}%
}

\pgfmathsetmacro{\azimuth}{100}
\pgfmathsetmacro{\elevation}{30}

\pgfmathsetmacro{\pointTheta}{60}
\pgfmathsetmacro{\pointPhi}{60}

\title{Implicit Inversion turns CLIP into a Decoder}

\author{%
  Antonio D'Orazio$^\dagger$\\
  \And
  Maria Rosaria Briglia$^\dagger$
  \And
  Donato Crisostomi$^{\star}$$^{\dagger}$\\
  \And
  Dario Loi$^\dagger$\\
  \And
  Emanuele Rodolà$^\star$\\
  \And
  Iacopo Masi$^\dagger$\\
}

\begin{document}

\maketitle
{\vspace{-20pt}
\centering
 \methodimagetag{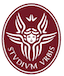}~~Sapienza University of Rome -  $^{\dagger}$~ \methodimagetag{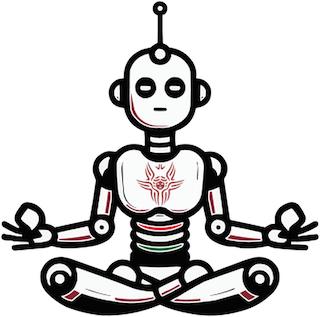} \href{https://omnai.di.uniroma1.it}{OmnAI Lab}~~ $^{\star}$~\methodimagetag{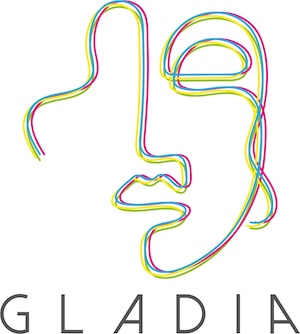} \href{https://gladia.di.uniroma1.it/}{GLADIA}
}{\centering
  %
  
%  \texttt{dorazio,briglia,crisostomi,rodola,masi@\{di.uniroma1.it\}} \\
}
\begin{figure}[ht]
  \centering
  %\vspace{-10pt}  
  \includegraphics[width=\textwidth]{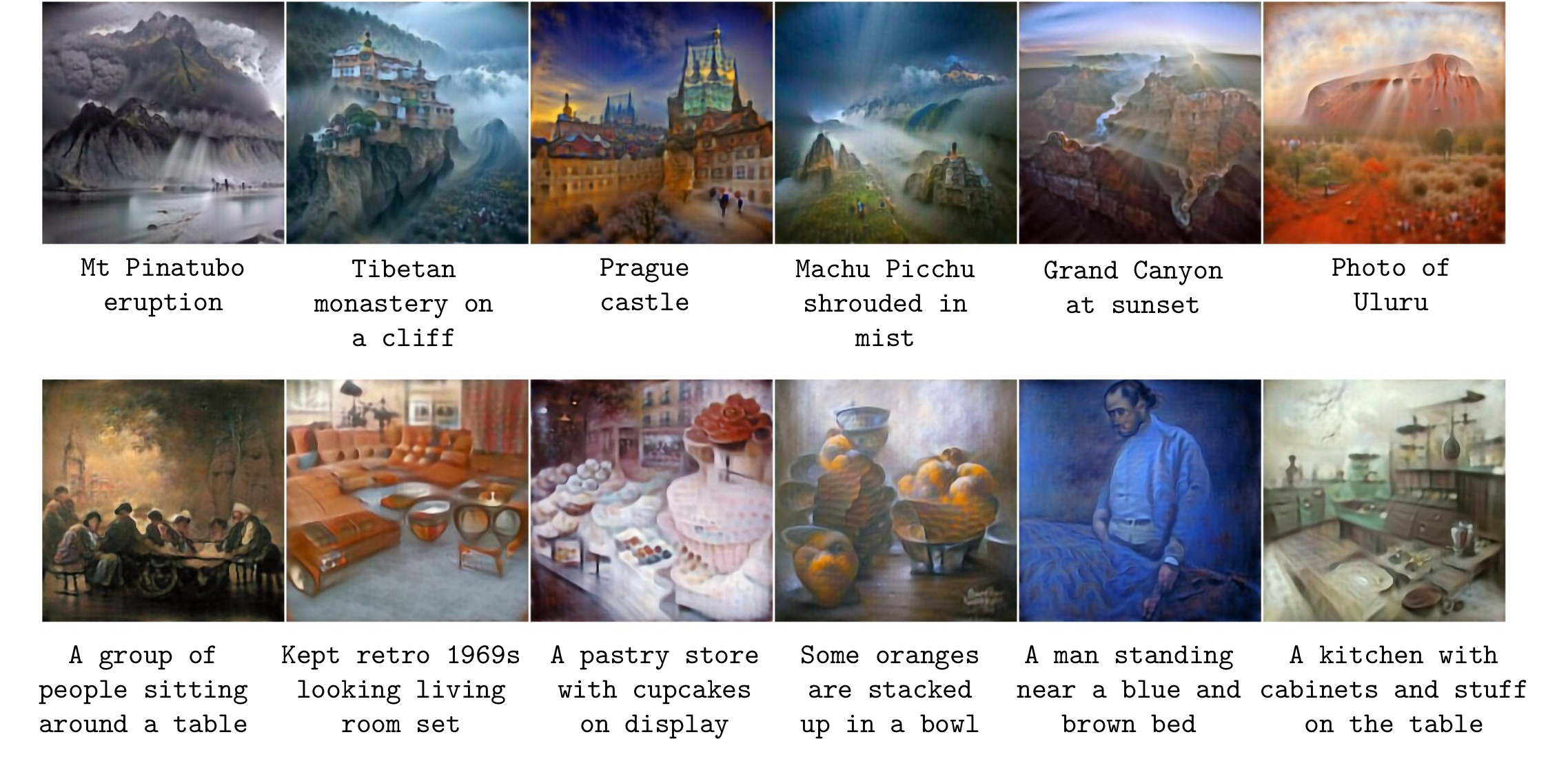}
  \label{fig:teaser}
    \caption{\small{\textbf{Decoder-free text-to-image synthesis.}} CLIP$^{-1}$ inverts CLIP's image encoder using implicit neural representations, enabling text-to-image synthesis \emph{without any fine-tuning or dedicated generative decoder}. All samples are generated with CLIP ViT-B/32~\cite{radford2021learning}, with \emph{top} rows showing generic scene prompts and \emph{bottom} rows illustrating complex captions from MS-COCO~\cite{lin2014microsoft}.
}
  \label{fig:teaser}
\end{figure}

\input{sec/00_abstract.tex}

\input{sec/01_intro.tex}

\input{sec/02_method.tex}

\input{sec/03_experiments.tex}

\input{sec/04_related_work.tex}

\input{sec/06_conclusions.tex}

{
  \small
  \bibliographystyle{ieeenat_fullname}
  \bibliography{main}
}

\renewcommand{\thefigure}{\Alph{figure}}
\input{sec/A_supmats}

% \newpage
% \input{sec/99_checklist.tex}

\end{document}

%% file: don_macros.tex
\definecolor{forestGreen}{RGB}{34, 139, 34}
\definecolor{firebrick}{RGB}{178, 34, 34}

\newcommand{\negxmark}{\large\color{firebrick}{×}}
\newcommand{\poscheckmark}{\large\color{forestGreen}{\checkmark}}

%% file: sec/00_abstract.tex
\begin{abstract}
    CLIP is a discriminative model trained to align images and text in a shared embedding space. Due to its multimodal structure, it serves as the backbone of many generative pipelines, where a decoder is trained to map from the shared space back to images. In this work, we show that image synthesis is nevertheless possible using CLIP alone---without any decoder, training, or fine-tuning. Our approach optimizes a frequency-aware implicit neural representation that encourages coarse-to-fine generation by stratifying frequencies across network layers. To stabilize this inverse mapping, we introduce adversarially robust initialization, a lightweight Orthogonal Procrustes projection to align local text and image embeddings, and a blending loss that anchors outputs to natural image statistics. Without altering CLIP’s weights, this framework unlocks capabilities such as text-to-image generation, style transfer, and image reconstruction. These findings suggest that discriminative models may hold untapped generative potential, hidden in plain sight. 
  \end{abstract}

%% file: sec/01_intro.tex
\section{Introduction} \label{sec:intro}
%
% Broader context
%
Text-to-image generation has progressed from a challenging research problem to a widely accessible technology, with recent models achieving photorealistic results and demonstrating remarkable creative abilities~ \cite{betker2023improving,saharia2022photorealistic,nichol2022glide, chang2023muse}.
Among these, latent diffusion models~\cite{rombach2022high} currently define the state-of-the-art by relying on an encoder–decoder architecture, where the encoder maps the input text into a latent representation, and the decoder reconstructs an image from it. Foundational vision-language models like CLIP \cite{radford2021learning} are frequently utilized as text encoders~\cite{betker2023improving,wang2022clip,tao2023galipgenerativeadversarialclips}, whereas the decoder is typically a diffusion model -- by far the most computationally demanding stage of the pipeline.

% Context
In this paper, \emph{we show that CLIP alone can perform text‑to‑image generation, without requiring a dedicated decoder.} 
We achieve this by inverting the CLIP vision encoder: rather than mapping an image to its latent embedding, we start from a CLIP embedding and reconstruct the corresponding image. Although prior work has attempted this through direct pixel space optimization~\cite{kazemi2024learn,ganz2023clipaggeneratorfreetexttoimagegeneration, ganz2024texttoimagegenerationenergybasedclip}, such approaches produce low-quality output with visible artifacts~\cite{kazemi2024learn} or require additional training~\cite{ganz2023clipaggeneratorfreetexttoimagegeneration, ganz2024texttoimagegenerationenergybasedclip}.
In contrast, we introduce CLIP$^{-1}$, a \emph{decoder-free and training-free} solution based on Implicit Neural Representations (INRs) \cite{liu2024finer}. \cref{fig:teaser} offers our results on generic scenes prompts and longer, more complex prompts from MS-COCO~\cite{lin2014microsoft}.
% Method overview
In practice, we first retrieve a seed, low-frequency INR whose caption is most similar to the prompt. Because this INR was pre‑trained on a blurred version of its image with Adversarial Weight Perturbations (AWP)~\cite{wu2020adversarial}, its low‑frequency weights are stable and provide a robust anchor.  
We then refine the INR layer by layer: a peaked learning‑rate schedule moves from low‑ to high‑frequency layers, progressively adding details in a coarse-to-fine manner. During refinement, we encourage on-manifold solutions by employing simple cosine losses against a style prompt and a small set of retrieved natural images, eventually resulting in more aesthetic generations. 
Avoiding direct pixel-space optimization, our approach naturally avoids structural artifacts and elegantly results in a coarse-to-fine generation akin to that of diffusion models. %With that being said, at its current stage, the method does not offer an alternative to diffusion models yet we believe our method bridges the gap between discriminative and generative models and can be an useful tool for understanding CLIP.

% Procrustes / other contributions
Despite the stability of the INR backbone, inversion still suffers from CLIP's modality gap \cite{liang2022mind, mistretta2025cross, zhang2024connect}: text and image embeddings fall on slightly offset sub‑manifolds, making the raw prompt embedding an unreliable target.
We bridge this gap by seeking an optimal orthogonal transformation via Procrustes analysis~\cite{procrustes, maiorca2023latent}, using the $k$ nearest caption–image pairs to align the prompt embedding with the image manifold and produce a well‑conditioned target for optimization.

% Evaluation
We benchmark our decoder‑free, training‑free pipeline on 10k MS‑COCO~\cite{lin2014microsoft} captions, reporting Fréchet Inception Distance (FID)~\cite{heusel2017gans}, Inception Score (IS)~\cite{salimans2016improved}, and CLIPSIM metric~\cite{hessel2021clipscore}. Compared with DAS~\cite{fort2025direct}, a concurrent decoder-free and training-free approach, our method achieves half the FID while almost doubling IS, producing noticeably crisper, more faithful images. The very same frozen model transfers zero‑shot to image reconstruction, controlled image edits, and neural style transfer, confirming the versatility of the INR backbone. Ablations show that \textit{(\romannumeral 1)} adversarially robust INRs boost quality, \textit{(\romannumeral 2)} Procrustes alignment yields sharper, more semantically aligned images, and \textit{(\romannumeral 3)} frequency‑steered optimization suppresses high‑frequency artifacts.

%% AT DONATO: ricordati di definire CLIP alla meno 1!!!

% Contribution list
In summary, our contribution is 3-fold.
\begin{enumerate}[itemsep=2pt, leftmargin=*]
    \item \textbf{Decoder–free, training–free CLIP inversion.}
    We repurpose a \emph{frozen} CLIP as a text‑to‑image generator by optimizing a \textbf{frequency‑aware Implicit Neural Representation} instead of generating images directly. Our pipeline \textit{(\romannumeral 1)} retrieves an adversarial‑robust, blur‑initialized INR that anchors low-frequency content, and \textit{(\romannumeral 2)} refines its weights with a coarse‑to‑fine, layer‑wise schedule guided solely by CLIP losses—eliminating the need for external decoders or re‑training.
    \item \textbf{Modality‑gap reduction with orthogonal Procrustes.} 
    We bridge the mismatch between CLIP’s text and image sub‑manifolds with a lightweight orthogonal transformation on the $k$ nearest caption–image pairs, projecting the prompt embedding into the image manifold to stabilize inversion.
    \item \textbf{Extensive empirical validation.}
    We extensively evaluate our pipeline, outperforming comparable prior work while demonstrating zero‑shot transfer to image reconstruction, controlled edits, and neural style transfer, with ablations confirming the benefit of each component.
\end{enumerate}
We publicly release all code and models to foster future research\footnote{Code is available at \href{https://github.com/omnai-lab/implicit-inversion}{github.com/omnai-lab/implicit-inversion}.}.

%% file: sec/02_method.tex
\section{Method}
% %%%%%%%%%%%%%%%%%%%%%%%%%%%%%%%%%%%%%%%%%%%%%%%%%%%%%%%%%%%%%
\begin{figure}[h]
  \begin{center}
    \begin{adjustbox}{max width=0.98\textwidth}
      \includegraphics[width=\linewidth]{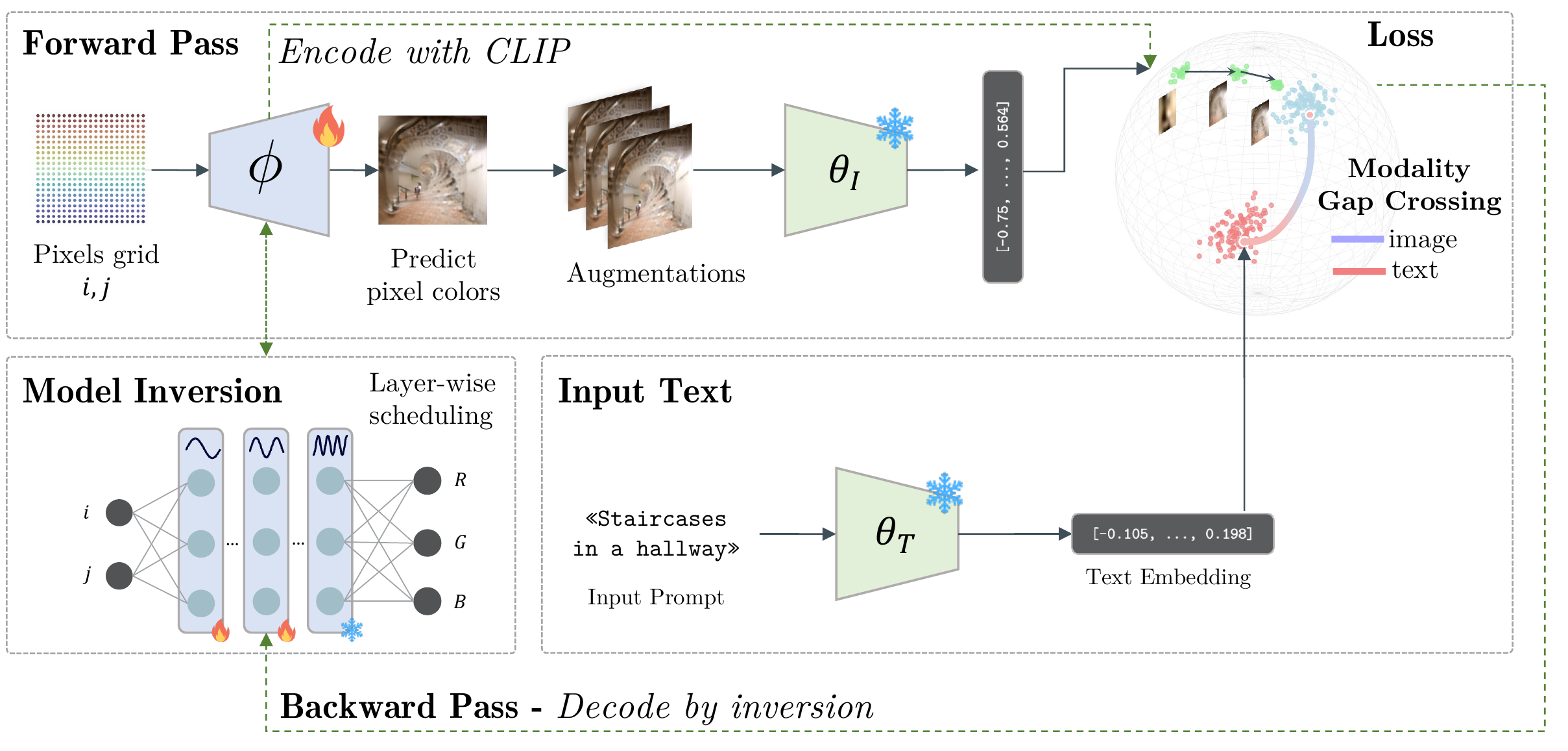}
    \end{adjustbox}
  \end{center}
  \caption{\small{\textbf{CLIP$^{-1}$ text-to-image inversion pipeline}}. The image is initially represented using an Implicit Neural Representation (INR) $f_\phi(i,j)$, optimizing the INR weights to match an input text prompt. The starting point for the inversion is a robust INR trained with Adversarial Weight Perturbation (AWP). The optimization updates the INR layer-wise, so that the embedding of its rendering aligns with the input text prompt embedding. The procedure includes image augmentations and CLIP embeddings averaging and projection onto the unit embedding sphere. In order to align text and image embeddings, we also apply an orthogonal Procrustes transformation to address the modality gap present in CLIP.
  }
  \label{fig:pipeline}
\end{figure}
% %%%%%%%%%%%%%%%%%%%%%%%%%%%%%%%%%%%%%%%%%%%%%%%%%%%%%%%%%%%%%
%
Given an input text prompt, our approach aims to generate images by inverting the corresponding CLIP embeddings.
The pipeline consists of three main stages:
\textit{(\romannumeral 1)} a \emph{data preparation} step (performed once offline),
\textit{(\romannumeral 2)} an \emph{initialization} step leveraging preprocessed data to retrieve a suitable starting point,
\textit{(\romannumeral 3)} an \emph{optimization} procedure refining the initial image to match the target text.

\subsection{Preliminaries}
\minisection{Implicit Neural Representations}
INRs represent images as functions that map spatial coordinates $(i, j)$ to RGB values $f_\phi(i, j) = (\text{r}, \text{g}, \text{b})$, where $\phi$ are the parameters of a neural network. The architecture is typically a Multilayer Perceptron (MLP), using positional encoding or frequency-aware activations for compact and efficient representation, crucial for image synthesis via iterative optimization. We adopt FINER~\cite{liu2024finer}, which enhances fine-detail modeling using variable periodic activations. It is based on SIREN~\cite{sitzmann2020implicit}, which uses the fixed-frequency activation function $\mathbf{z}_i = \sin\left(\omega(\mathbf{W}_i\mathbf{z}_{i-1}+\mathbf{b}_i)\right)$. FINER improves adaptability by introducing an additional coefficient:
\begin{align}
  \begin{gathered}
    \mathbf{z}_i = \sin\big( \omega \alpha_i ( \mathbf{W}_i \mathbf{z}_{i-1} + \mathbf{b}_i) \big), \quad
    \alpha_i = \left| \mathbf{W}_i \mathbf{z}_{i-1} + \mathbf{b}_i \right| + 1\,,
  \end{gathered}
\end{align}
where $\alpha_i$ dynamically adjusts local frequency based on input magnitude. We also leverage FINER’s bias initialization, which stratifies frequencies across layers, capturing low frequencies early and high frequencies deeper, leading to better convergence and reconstruction. See \cref{fig:ablation} for an illustration.

\subsection{Data preparation}\label{sec:preprocessing}
The data preparation stage is performed offline once and sets up the foundation for inversion initialization. In practice, we used images from LAION Aesthetics (a subset of LAION-5B~\cite{schuhmann2022laion}). This data serves two purposes: \textit{(\romannumeral 1)} training INRs for initializing text-to-image optimization and \textit{(\romannumeral 2)} computing natural image CLIP embeddings to act as anchor points in the latent space.

Each image is first blurred using a Gaussian filter to suppress high frequencies, providing a smoother starting point for inversion. An INR is then trained to reconstruct the blurred image, and its weights are stored.
For the $i$-th image, the predicted pixel values from the INR are $f_{\phi^i}$, with weights $\phi^i$. The INR images are then encoded via CLIP to obtain visual embeddings $\clipi{f_{\phi^i}}$. The CLIP text embeddings $\clipt{\prompt^i}$ of the corresponding captions are also stored. This results in a dataset $\mathcal{D}= \{\clipi{f_{\phi^i}},f_{\phi^i},\clipt{\prompt^i}\}$, containing CLIP image embeddings, INRs, and CLIP text embeddings for each training sample.

\minisection{Robust INR initialization}
The INR weights $\phi^i$ can be viewed as compact embeddings of the $i$-th image -- each uniquely capturing its content. However, small perturbations in these weights can significantly alter the reconstructed image, making them sensitive and potentially unstable for downstream tasks.

To address this, we propose a training method that improves INR weights robustness  by incorporating adversarial perturbations during training. Given an INR with weights $\phi$, we define an adversarial weight perturbation (AWP) $\Delta \phi \in \Omega$, where $\Omega$ bounds the perturbation range. The training objective is to make the model resilient to such perturbations by solving the following min-max optimization:
 \begin{equation}
  %\min_{\phi} \max_{\Delta{\phi} \in \Omega} \rho({\phi} + \Delta{\phi}) \rightarrow 
  \min_{\phi} \max_{\Delta{\phi} \in \Omega}\Loss\left(f_{{\phi} + \Delta{\phi}},\text{blur}(\img)\right)\,,
  \label{eq:min-max-eq}
 \end{equation} 
where $f_{{\phi} + \Delta{\phi}}$ is the perturbed INR output and $\Loss$ is the reconstruction loss function w.r.t. the blurred target image $\img$. 
Unlike standard adversarial training in the input space~\cite{wu2020adversarial}, our approach perturbs only the model weights. The perturbation set is constrained by a relative norm bound $\Omega = \{ \Delta : \norm{\Delta} \leq \gamma \norm{{\phi}} \}$, with $\gamma$ controlling the allowed perturbation magnitude. 
This robust training is applied only once, during the construction of the initial INR at the inversion step $n=0$. It ensures that early optimization steps do not cause the weights to drift too far from the frequency content of the initialization, improving stability during inversion. The supplementary material details the training procedure using the AWP algorithm.

\subsection{Initialization and modality gap handling}

\minisection{Inversion initialization}
In text-to-image generation, the process begins with a text prompt $\prompt$, which is encoded using the CLIP text encoder to obtain $\embtext = \clipt{\prompt} \in \mathbb{R}^d$. To initialize the inversion, we search our dataset $\mathcal{D}$ for the INR whose associated caption has the highest cosine similarity to $\embtext$. This serves as the starting point for the optimization.

\minisection{Bridging the modality gap}
CLIP aligns text and image embeddings globally, projecting them onto the unit sphere. However, local differences between modalities persist: text embeddings tend to encode abstract semantics, while image embeddings reflect concrete visual features.
Directly optimizing an image to match a text embedding often causes artifacts or \textit{textual hallucinations} -- the model overfits to abstract concepts and produces unrealistic visuals~\cite{NEURIPS2022_702f4db7}.

To address this, we learn a local transformation to align text and image embeddings more precisely. We retrieve the $k$ nearest neighbors of the input text embedding $\embtext$ from our dataset $\mathcal{D}$, forming two matrices: $\mbf{E}_T \in \mathbb{R}^{d \times k}$, encoding the $k$ closest text embeddings, and $\mbf{E}_I \in \mathbb{R}^{d \times k}$ encoding their corresponding image embeddings.
Solving the orthogonal Procrustes problem we find an orthogonal matrix $\mbf{R}$ that best aligns these two sets:
\begin{align}
  \begin{gathered}
    \min_{\mbf{R}} ||\mbf{R}\mbf{E}_T-\mbf{E}_I||_F \qquad \text{s.t.}\  \mbf{R}^\top\mbf{R}=\mbf{I}\,,
    \label{eq:ortoproc}
  \end{gathered}
\end{align}
where $\|\cdot\|_F$ is the Frobenius norm. This procedure is performed individually for each input prompt. The resulting orthogonal matrix $\mbf{R}$ aligns the local structure of text embeddings with that of image embeddings. 
We then transform the input text embedding into the image modality as $\embtexttoim = \mbf{R}\embtext$, which becomes the target embedding for the CLIP inversion process.

\subsection{Inverting CLIP with Implicit Neural Representations}
\label{sec:invertingclip}
\minisection{Text-To-Image via CLIP Inversion}
Our pipeline, shown in \cref{fig:pipeline}, inverts a CLIP text embedding to synthesize an image using an INR. Given a text prompt $\prompt$, we obtain the projected image-space embedding $\embtexttoim = \mbf{R}\clipt\prompt$. The INR $f_\phi$ is optimized so that its output image matches $\embtexttoim$ when passed through the frozen CLIP image encoder $\clipisingle$. This is formalized as:
\begin{equation}
  \phi = \arg\min_\phi \Loss(\clipi {f_\phi}, \embtexttoim) \quad \text{where}  \quad \embtexttoim = \mbf{R}\clipt{\prompt}.
\end{equation}
Here, $\Loss$ is the cosine distance, and gradients flow from CLIP back to the INR parameters $\phi$.

\begin{wrapfigure}[16]{R}{0.35\textwidth}
  \vspace{-0.75cm}
  \centering
  \resizebox{1\linewidth}{!}{
    \includegraphics[width=0.75\linewidth]{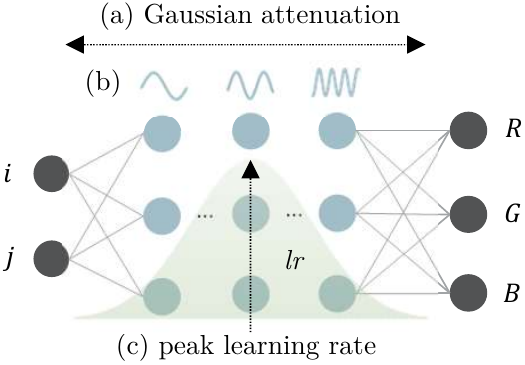}
  }
  \caption{
    \textbf{\small{Gaussian Scheduling.}} Each layer represents a frequency interval \textit{(b)} for $f_\phi$. The learning rate is centered on a specific layer and gradually shifted \textit{(c)}, decreasing with a Gaussian attenuation across neighboring layers \textit{(a)}. 
  }
  \label{fig:gausslr}
\end{wrapfigure}

\minisection{Layer-wise frequency optimization}
Rather than optimizing pixel values directly, we update the INR weights, leveraging FINER's property that network layers correspond to different frequency bands. The INR is structured as an $L$-layer MLP, with each layer representing a specific frequency range.
To guide the optimization process, we apply Gaussian learning rate scheduling: at each iteration, we focus the optimization on a specific layer by assigning it a peak learning rate, while attenuating the rates of neighboring layers according to a Gaussian curve (see \cref{fig:gausslr}). This helps reconstruct coarse features before fine details, improving stability and fidelity.

\minisection{Augmentations for stable optimization}
Inspired by CLIPDraw~\cite{10.5555/3600270.3600646} and CLIPAG~\cite{ganz2023clipaggeneratorfreetexttoimagegeneration}, we apply image augmentations such as color shifts, scaling, and shearing, while optimizing. Each augmented image is encoded with CLIP, and their embeddings are averaged and projected back onto the unit sphere: $\mbf{e}_i^\star = \frac{1}{n} \sum_{k=1}^{n} \clipi{\text{augment}(f_{\phi^k})}$, strongly enforcing synthesized image robustness to visual distortions.

\minisection{Blending natural image priors} 
To further guide generation toward realistic outputs, we incorporate information from natural images. For a given prompt $\prompt$, we retrieve the $k$ most similar image embeddings from a reference dataset using cosine similarity. These are linearly combined into a blended target embedding $\mbf{e}^\star_{img}$, with weights given by the softmax of the similarity scores. A blending loss $\Loss_{\text{blend}}$ then encourages the output embedding $\mbf{e}_i^\star$ to remain close to the manifold.

\minisection{Final optimization formulation} 
The complete CLIP$^{-1}$ optimization pipeline updates the INR parameters $\phi$ to generate realistic images, leveraging both augmented embeddings and natural image priors, with no CLIP retraining or modification. The full procedure is:
% \vspace{3pt}
\begin{align}
  \begin{split}
    \begin{cases}
      \begin{array}{l}
        \vspace{6pt}
        % INR Image forward pass
        \text{(a)} \quad \mbf{e}_i^\star = \frac{1}{n} \sum_{k=1}^{n} \clipi{\text{augment}(f_{\phi^k})} \\                                                         
        % CLIP Inversion
        \text{(b)} \quad \phi_0 = \min_{\phi} \max_{\Delta{\phi} \in \Omega} \Loss\big(f_{{\phi} + \Delta{\phi}},\text{blur}(\img)\big) \\ %& (c)
        \text{(c)} \quad \phi_{n} = \phi_{n-1} -  \nabla_{\phi} \Big[\Loss(\mbf{e}_i^\star, \embtexttoim) + \beta \Loss_{blend}\left( \mbf{e}_i^\star,\mbf{e}^\star_{img}\right) \Big] 
      \end{array}
    \end{cases}
  \end{split}
  \label{equ:opt-full}
\end{align}
Step (a) computes the CLIP embedding from the augmented INR outputs; step (b) initializes the INR with adversarial weight perturbations to enhance robustness, and (c) updates the INR weights via backpropagation using both alignment and blending losses.

%% file: sec/03_experiments.tex
\section{Experiments}
\label{sec:experiments}
We evaluate our decoder-free, training-free inversion pipeline across a range of tasks. We begin with text-to-image generation on \dataset{MS-COCO}~\cite{lin2014microsoft}, comparing against both standard generative models and prior inversion-based approaches (\S\ref{sec:texttoimage}). Next, we demonstrate the generality of our method through zero-shot transfer to downstream tasks, including reconstruction, controlled modification, and style transfer (\S\ref{subsec:downstream-tasks}). Lastly, we quantify the effect of each component through an ablation study (\S\ref{subsec:ablation}).

\input{tables/main-results}

\begin{figure}[t]
  \centering
  \begin{tabular}{@{\,}c@{\,}c@{\,}c@{\,}c@{\,}c@{\,}c@{\,}c@{\,}c@{\,}}
    \scriptsize \parbox[c]{0.12\linewidth}{\centering \texttt{CLIPAG                                                                                   \\ \cite{ganz2023clipaggeneratorfreetexttoimagegeneration}}} &
    \scriptsize \parbox[c]{0.12\linewidth}{\centering \texttt{CLIP-JEM \\\cite{ganz2024texttoimagegenerationenergybasedclip}}} &
    \scriptsize \parbox[c]{0.12\linewidth}{\centering \texttt{DAS                                                                                      \\ \cite{fort2025direct}}} &
    \scriptsize \parbox[c]{0.12\linewidth}{\centering \texttt{CLIP-Inv \cite{kazemi2024learn}}}                              &
    \scriptsize \parbox[c]{0.12\linewidth}{\centering \tbf{CLIP$^{-1}$}                                                                                \\\texttt{(ViT-B/32})} &
    \scriptsize \parbox[c]{0.12\linewidth}{\centering \tbf{CLIP$^{-1}$}                                                                                \\\texttt{(RESNET)}} &
    \scriptsize \parbox[c]{0.12\linewidth}{\centering \tbf{CLIP$^{-1}$}                                                                                \\\texttt{w/\cite{ganz2023clipaggeneratorfreetexttoimagegeneration} }} &
    \scriptsize \parbox[c]{0.12\linewidth}{\centering \tbf{CLIP$^{-1}$}                                                                                \\\texttt{w/\cite{ganz2024texttoimagegenerationenergybasedclip} (XXL)}} \\

    \includegraphics[width=0.12\linewidth]{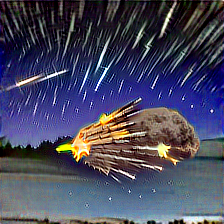}                                                  &
    \includegraphics[width=0.12\linewidth]{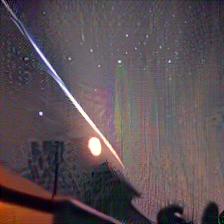}                                                 &
    \includegraphics[width=0.12\linewidth]{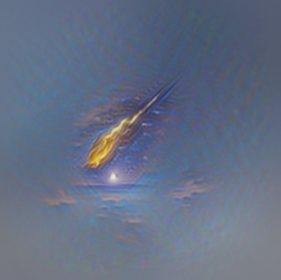}                                               &
    \includegraphics[width=0.12\linewidth]{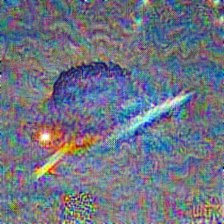}                                                 &
    \includegraphics[width=0.12\linewidth]{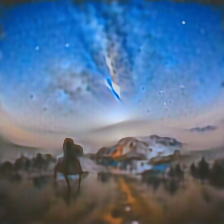}                                                    &
    \includegraphics[width=0.12\linewidth]{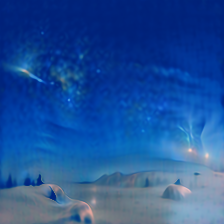}                                             &
    \includegraphics[width=0.12\linewidth]{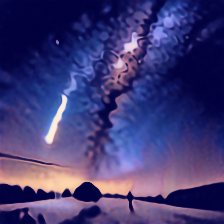}                                             &
    \includegraphics[width=0.12\linewidth]{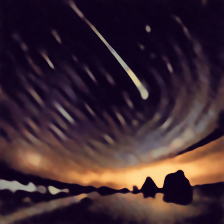}                                                                      \\
    \multicolumn{8}{c}{\scriptsize \texttt{\guillemotleft Meteor streaking through the night sky\guillemotright}}                                      \\
    \vspace{-5pt}                                                                                                                                       \\

    \includegraphics[width=0.12\linewidth]{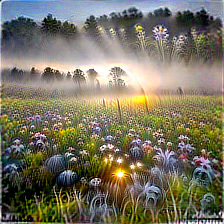}                                                  &
    \includegraphics[width=0.12\linewidth]{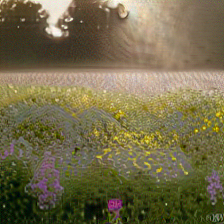}                                                 &
    \includegraphics[width=0.12\linewidth]{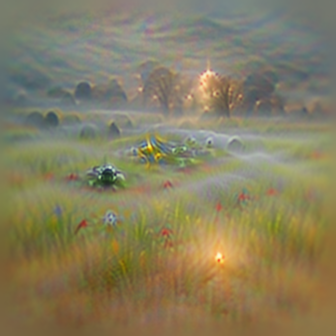}                                               &
    \includegraphics[width=0.12\linewidth]{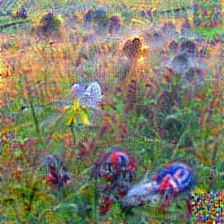}                                                 &
    \includegraphics[width=0.12\linewidth]{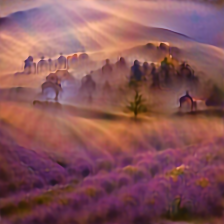}                                                    &
    \includegraphics[width=0.12\linewidth]{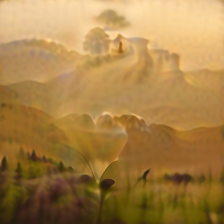}                                             &
    \includegraphics[width=0.12\linewidth]{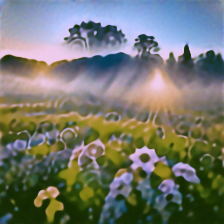}                                             &
    \includegraphics[width=0.12\linewidth]{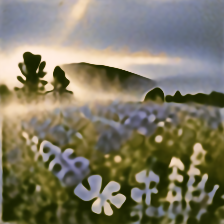}                                                                      \\
    \multicolumn{8}{c}{\scriptsize \texttt{\guillemotleft A mist-covered field at daybreak with wildflowers glistening in early rays.\guillemotright}} \\
  \end{tabular}
  \caption{\small{\textbf{Qualitative comparison}} of prior pixel-based methods against different \small{\textbf{CLIP$^{-1}$}} configurations. }
  \label{fig:qualitative-eval}
\end{figure}

\subsection{Generator-free text-to-image synthesis}
\minisection{Setting}
In text-to-image synthesis, the goal is to generate visually realistic images that are semantically aligned with a natural language description. To assess the visual fidelity of our method, we compute the Fréchet Inception Distance (FID)~\cite{heusel2017gans} and the Inception Score (IS)~\cite{salimans2016improved} over a subset of $10,000$ captions from \dataset{MS-COCO}~\cite{lin2014microsoft}. Since these metrics do not capture semantic alignment with the prompt, we also report CLIPSIM~\cite{hessel2021clipscore}, which measures the cosine similarity between the CLIP embeddings of generated images and their corresponding captions.
We compare our method against prior and concurrent decoder-free approaches, including those that require fine-tuning~\cite{ganz2023clipaggeneratorfreetexttoimagegeneration, ganz2023perceptually} and those that do not~\cite{fort2025direct, kazemi2024learn}. As our method requires no training or decoder, we regard tuning-free baselines as the most relevant points of comparison. For completeness, we also report results from state-of-the-art generative models that rely on both training and a dedicated decoder~\cite{betker2023improving, nichol2022glide, rombach2022high}. The implementation details can be found in the supplementary material.

\minisection{Results}
\Cref{tab:main_results} reports our results alongside model sizes and architectural requirements. Among decoder-free, training-free methods, CLIP$^{-1}$ achieves the lowest FID (72.5 vs.\ 161.8 for \model{DAS-ViT}~\cite{fort2025direct}) and highest IS (9.5 vs. 5.7), marking a substantial improvement in visual quality. Although diffusion-based models still attain lower FID, they require orders of magnitude more parameters and full training pipelines, whereas our method uses a frozen backbone and a lightweight INR.
Finally, our method achieves a CLIPSIM of 38.6, outperforming both training-free and fine-tuned baselines---except for CLIPInvert~\cite{kazemi2024learn}, whose higher CLIPSIM can be attributed to overfitting to the target embedding, as evidenced by its elevated FID and low IS. Taken together, these results indicate that Procrustes alignment and frequency-aware INR optimization effectively improve text-image consistency without modifying CLIP's weights.
\Cref{fig:qualitative-eval} further provides qualitative examples: compared to other training-free baselines, our generations exhibit fewer structural artifacts and sharper details, and visually approach the quality of tuned approaches~\cite{ganz2023clipaggeneratorfreetexttoimagegeneration, ganz2024texttoimagegenerationenergybasedclip}.
We also apply our pipeline in a plug-and-play fashion to tuned CLIP variants, such as CLIPAG~\cite{ganz2023clipaggeneratorfreetexttoimagegeneration} and CLIP-JEM~\cite{ganz2024texttoimagegenerationenergybasedclip}, showing broader compatibility of our method with discriminatively trained models.

\label{sec:texttoimage}

% \subsection{Image In-painting}

% \begin{itemize}
%     \item Viene istanziata una INR che fitta totalmente l'immagine
%     \item Viene croppata la parte su cui fare painting e sì fa detach della parte su cui non farlo
%     \item Si fa una nuova image $I_{detached} + I_{INR cropped}$
%     \item si ottimizza il crop con input prompt
% \end{itemize}
\begin{figure}[t]
  \centering
  \renewcommand{\arraystretch}{0.3}
  \vspace{16pt}
  \begin{subfigure}[t]{0.32\linewidth}
    \centering
    \resizebox{\linewidth}{!}{
      \begin{tabular}{@{}c@{\hspace{1pt}}c@{\hspace{1pt}}c@{}}
        \includegraphics[width=0.3\linewidth]{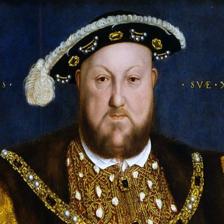}  &
        \includegraphics[width=0.3\linewidth]{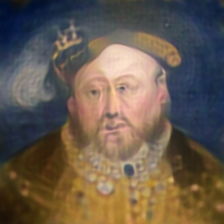}        &
        \includegraphics[width=0.3\linewidth]{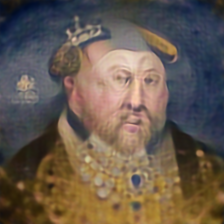}                                            \\
        \includegraphics[width=0.3\linewidth]{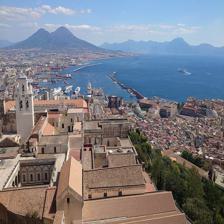} &
        \includegraphics[width=0.3\linewidth]{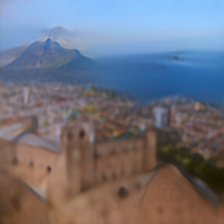}       &
        \includegraphics[width=0.3\linewidth]{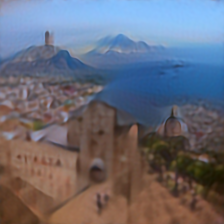}                                           \\
        \includegraphics[width=0.3\linewidth]{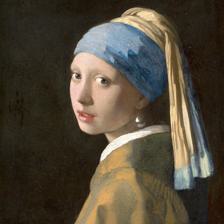} &
        \includegraphics[width=0.3\linewidth]{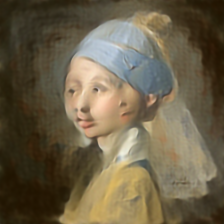}       &
        \includegraphics[width=0.3\linewidth]{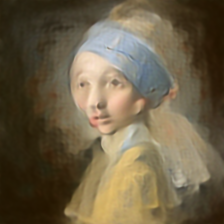}                                           \\
        \vspace{0.05em}                                                                               & \vspace{0.05em} & \vspace{0.05em} \\
        \scriptsize \texttt{Original}                                                                 &
        \scriptsize \texttt{Rec. 1}                                                                   &
        \scriptsize \texttt{Rec. 2}                                                                                                       \\
      \end{tabular}
    }
    \caption{\small Reconstruction}\label{fig:task-reconstruction} % Caption specifico per questa subfigure
    \vspace{5pt} % Spazio tra la tabella e il testo
  \end{subfigure}%
  \hfill
  \begin{subfigure}[t]{0.32\linewidth}
    \centering
    \resizebox{\linewidth}{!}{
      \begin{tabular}{@{}c@{\hspace{1pt}}c@{\hspace{1pt}}c@{}}
        \includegraphics[width=0.3\linewidth]{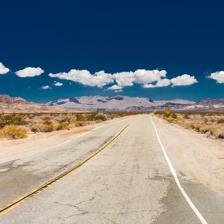} &
        \includegraphics[width=0.3\linewidth]{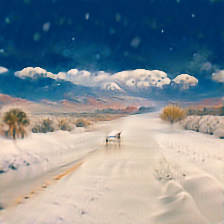}    &
        \includegraphics[width=0.3\linewidth]{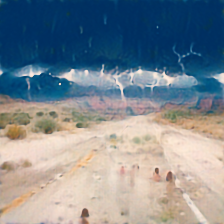}                                     \\
        \includegraphics[width=0.3\linewidth]{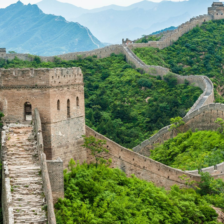}      &
        \includegraphics[width=0.3\linewidth]{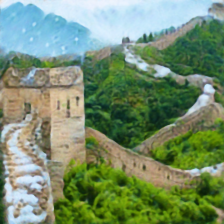} &
        \includegraphics[width=0.3\linewidth]{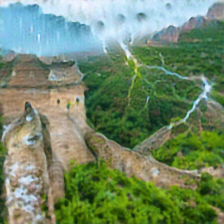}                                  \\
        \includegraphics[width=0.3\linewidth]{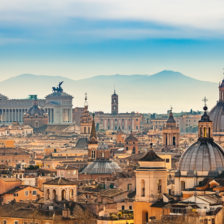}            &
        \includegraphics[width=0.3\linewidth]{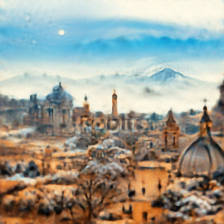}       &
        \includegraphics[width=0.3\linewidth]{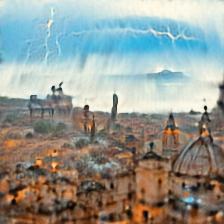}                                        \\
        \vspace{0.05em}                                                                               & \vspace{0.05em} & \vspace{0.05em} \\
        \scriptsize \texttt{Original}                                                                 &
        \scriptsize \texttt{Prompt 1}                                                                 &                                   % Peaceful snowy landscape
        \scriptsize \texttt{Prompt 2}                                                                                                     \\ % Torrential rainfall, lightning bolts
      \end{tabular}
    }
    \caption{\small Controlled Modification}\label{fig:task-modification} % Caption specifico per questa subfigure
    \vspace{5pt} % Spazio tra la tabella e il testo
  \end{subfigure}
  \begin{subfigure}[t]{0.32\linewidth}
    \centering
    \resizebox{\linewidth}{!}{
      \begin{tabular}{@{}c@{\hspace{1pt}}c@{\hspace{1pt}}c@{}}
        \includegraphics[width=0.3\linewidth]{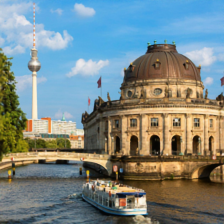}    &
        \includegraphics[width=0.3\linewidth]{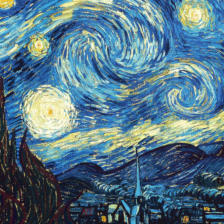}     &
        \includegraphics[width=0.3\linewidth]{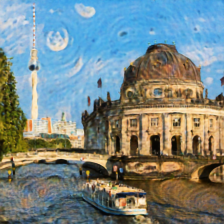}                                         \\
        \includegraphics[width=0.3\linewidth]{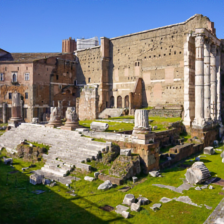}       &
        \includegraphics[width=0.3\linewidth]{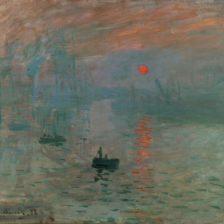}        &
        \includegraphics[width=0.3\linewidth]{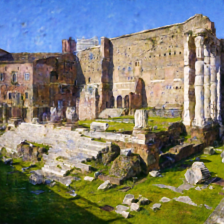}                                            \\
        \includegraphics[width=0.3\linewidth]{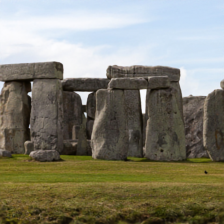} &
        \includegraphics[width=0.3\linewidth]{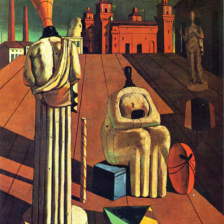}  &
        \includegraphics[width=0.3\linewidth]{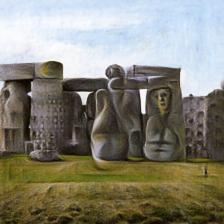}                                      \\
        \vspace{0.05em}                                                                             & \vspace{0.05em} & \vspace{0.05em} \\
        \scriptsize \texttt{Original}                                                               &
        \scriptsize \texttt{Reference}                                                              &
        \scriptsize \texttt{Result}                                                                                                     \\
      \end{tabular}
    }
    \caption{\small Neural Style Transfer}\label{fig:task-style-transfer} % Caption specifico per questa subfigure
    \vspace{5pt} % Spazio tra la tabella e il testo

  \end{subfigure}%
  \hfill

  \caption{\small{\tbf{Downstream Tasks.}} (a) Recreates the input image from its corresponding CLIP encoding. (b) Alters the input image based on a specified prompt. \footnotesize{\texttt{Prompt 1: \guillemotleft Snowy peaceful landscape\guillemotright}} \space; \footnotesize{\texttt{Prompt 2: \guillemotleft Torrential rainfall, lightning bolts\guillemotright}}. (c) Applies the visual style of a reference image to the input.
    }
  \label{fig:comparison_grids}
\end{figure}

\subsection{Zero-shot task generalization} \label{subsec:downstream-tasks}
To assess the versatility of our decoder-free approach, we explore several downstream tasks, demonstrating that the same inversion framework can successfully generate images in different settings without requiring task-specific modifications or additional optimization.

\minisection{Image reconstruction}
In this task, the goal is to reconstruct a given input image using our inversion pipeline. We treat the image as a target whose CLIP embedding is known, and optimize an INR to produce an output that matches this embedding. The INR is initialized from a blurred version of the image, and refined to align with the embedding of the full-resolution input, effectively operating as a decoder that recovers semantic content from latent space.
Unlike text-to-image generation, this task provides a well-defined ground truth and serves as a controlled setting to evaluate inversion precision. \Cref{fig:task-reconstruction} shows qualitative results on both artistic and photographic inputs. Across all examples, high-level semantic content, such as facial identity or scene composition, is consistently preserved. Fine-grained spatial details, especially in structured regions like faces or buildings, are approximated with some distortion or shift, reflecting the inherent ambiguity of CLIP’s embedding space.

\input{tables/ms-coco.tex}
\minisection{Controlled image modification}
In this task, the goal is to modify an input image according to a natural language prompt that specifies a targeted change in content or style. The image is first encoded via an INR fitted to its original form. A text prompt is then provided to guide the modification (e.g., \textit{``snowy landscape''} or \textit{``torrential rainfall''}).
The INR is optimized to align the CLIP embedding of the generated image with that of the prompt, while starting from the original image representation. This setup encourages localized, semantically consistent transformations without disrupting the broader structure or identity of the scene.
\Cref{fig:task-modification} shows three examples for the task. In each row, the left-most column is the original image; the next two columns show the edits for ``snow'' and ``storm''. The road, the Great Wall, and the city keep their geometry and colour palette, while only the requested weather effects (snow cover, rain streaks, lightning) are added. This confirms that CLIP$^{-1}$ can act as a prompt-driven image editor, producing targeted edits without explicit masks or additional training.

\minisection{Neural style transfer}
For style transfer we supply two images: a \emph{content} photo and a \emph{style} reference. The content image is represented by an INR initialized to exactly reproduce the original photo; the style image is fed only through the frozen CLIP encoder. Optimization minimizes a weighted sum of two CLIP-based losses:
\textit{(\romannumeral 1)} a \textit{style loss} that pulls the INR's embedding towards that of the reference painting, and
\textit{(\romannumeral 2)} a \textit{content loss} (weight 0.5) that keeps the embedding close to the original photo.
\Cref{fig:task-style-transfer} illustrates the outcome. In each case, the brush-stroke texture and overall palette of the reference painting are transferred, while object layout and scene geometry remain intact. The method therefore separates appearance from semantics without hand-crafted losses or additional training, indicating that the inversion pipeline can exploit CLIP's latent space to disentangle style from content.
\subsection{Ablation study} \label{subsec:ablation}

\begin{figure}[t]
  \centering
  \renewcommand{\arraystretch}{0.3}
  \begin{subfigure}[t]{0.48\linewidth}
  \vspace{-1.5cm}
  \centering
  \renewcommand\arraystretch{1.2}
  \label{tab:abl-qual}
  \resizebox{\textwidth}{!}{%
    \begin{tabular}{clccc}
      \toprule
      & Variant              & FID$\;\downarrow$ & CLIPSIM$\;\uparrow$ & IS$\;\uparrow$ \\
      \cdashlinelr{1-5}
      \textit{\romannumeral 1.} & \textbf{CLIP$^{-1}$} & \textbf{107.1}    & 38.8                & 7.7            \\
      \textit{\romannumeral 2.} & w/o Freq.\ Opt  (F.O.)       & 185.1             & 30.5                & 7.8            \\
      \textit{\romannumeral 3.} & w/o AWP               & 121.0             & 43.0                & 7.3            \\
      \textit{\romannumeral 4.} & w/o F.O. w/o Proc.             & 111.3             & 46.4                & \textbf{9.1}   \\
      \textit{\romannumeral 5.} & w/o F.O. w/o Blending Loss     & 119.7             & \textbf{49.5}       & 7.9            \\
      \bottomrule
    \end{tabular}}%
    \subcaption{}
  \end{subfigure}%
  \hfill
  \begin{subfigure}[t]{0.48\linewidth}
  \centering
  \resizebox{0.9\linewidth}{!}{
    % \begin{tabular}{@{}c@{\hspace{1pt}}c@{\hspace{1pt}}c@{}}
    \begin{tabular}{c@{\hspace{1pt}}c@{\hspace{1pt}}c@{\hspace{1pt}}c@{\hspace{1pt}}c}
      \includegraphics[width=0.15\linewidth]{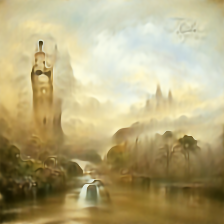}      &
      \includegraphics[width=0.15\linewidth]{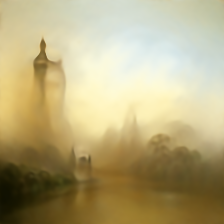}      &
      \includegraphics[width=0.15\linewidth]{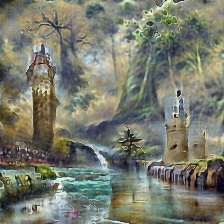}      &
      \includegraphics[width=0.15\linewidth]{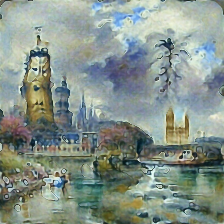} &
      \includegraphics[width=0.15\linewidth]{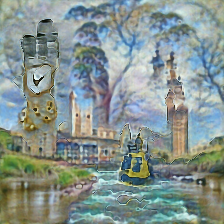}                                  \\
      \vspace{-3pt} \\
      
      \includegraphics[width=0.15\linewidth]{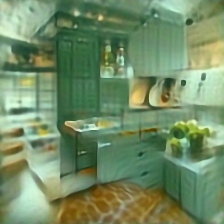}      &
      \includegraphics[width=0.15\linewidth]{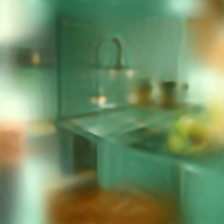}      &
      \includegraphics[width=0.15\linewidth]{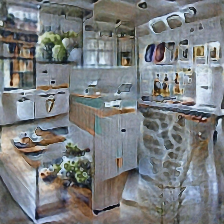}      &
      \includegraphics[width=0.15\linewidth]{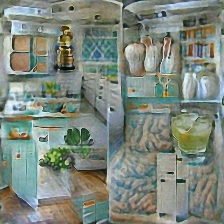} &
      \includegraphics[width=0.15\linewidth]{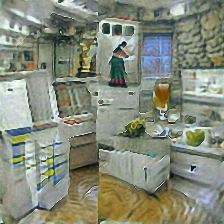}                                  \\
      \vspace{-3pt} \\
      \vspace{-3pt} \\
      \includegraphics[width=0.15\linewidth]{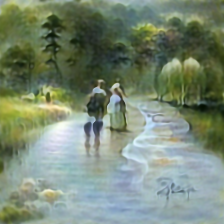}      &
      \includegraphics[width=0.15\linewidth]{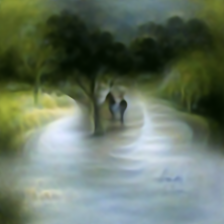}      &
      \includegraphics[width=0.15\linewidth]{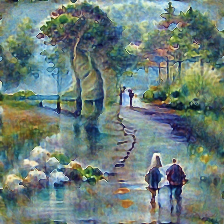}      &
      \includegraphics[width=0.15\linewidth]{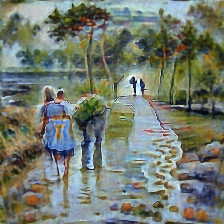} &
      \includegraphics[width=0.15\linewidth]{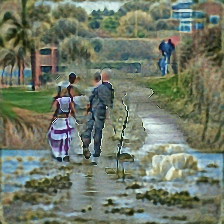}                                         \\
      
      \scriptsize \textit{\romannumeral 1.}                                                                 &
      \scriptsize \textit{\romannumeral 2.}                                                                 &                                   % Peaceful snowy landscape
      \scriptsize \textit{\romannumeral 3.}                                                                 &                                   % Peaceful snowy landscape
      \scriptsize \textit{\romannumeral 4.}                                                                 &                                   % Peaceful snowy landscape
      \scriptsize \textit{\romannumeral 5.}                                                                                                     \\ % Torrential rainfall, lightning bolts
    \end{tabular}
  }\subcaption{}\label{fig:qualitative-abl}

  \end{subfigure}

  \caption{\small{\textbf{Quantitative ablation study.}} (a) Results on 1{,}000 MS-COCO captions. Lower FID is better; higher CLIPSIM and IS indicate better performance. (b) Samples for each case. Rows share the same prompt; columns show: -- \textit{\romannumeral 1.} full model -- \textit{\romannumeral 2.}  frequency scheduling -- \textit{\romannumeral 3.} AWP -- \textit{\romannumeral 4.} F.O. + Procrustes -- \textit{\romannumeral 5.} F.O. + blending loss.}
  \label{fig:quantitative-ablation}
\end{figure}
We now perform a controlled ablation over the four key components: layerwise frequency scheduling, adversarial weight perturbation (AWP), orthogonal Procrustes alignment, and the natural-image blending loss; the corresponding ablated variants are referred to as \textit{(\romannumeral 1)}, \textit{(\romannumeral 2)}, \textit{(\romannumeral 3)}, \textit{(\romannumeral 4)}. We run every variant in the same $1000$ captions from \dataset{MS-COCO} and report FID~\cite{heusel2017gans}, CLIPSIM~\cite{hessel2021clipscore}, and IS~\cite{salimans2016improved}; Better FID/IS and higher CLIPSIM show better perceptual realism and stronger text–image agreement, respectively. In parallel, we visualize representative generations so that the numerical shifts can be linked to visual outcomes. Additional results can be found in the supplementary material.

The ablation shows that each proposed component plays a distinct, complementary role. When the layer-wise learning-rate schedule is removed \textit{(\romannumeral 2)} the INR is forced to optimize all frequency bands simultaneously; high-frequency layers overfit first, so fine textures emerge before the coarse layout has stabilized; the premature detail introduces stripe-like artifacts and drives FID to its worst value. Dropping AWP \textit{(\romannumeral 3)} preserves the coarse-to-fine dynamic but does not constrain weights to remain on the manifold defined by the robust anchor, introducing neural artifacts; this allows the unconstrained result to align more closely with the caption, increasing  CLIPSIM at the expense of realism (FID $\uparrow$). Similarly, eliminating the orthogonal Procrustes projection \textit{(\romannumeral 4)} pushes the optimization toward the raw text embedding, i.e. slightly outside the image sub-manifold: CLIP rewards the closer alignment (CLIPSIM $\uparrow$), but the outputs become noticeably busier, with sharper outlines, cluttered details, and occasional duplicated elements.
Finally, disabling the blending loss \textit{(\romannumeral 5)} stops the optimizer from referencing real-photo statistics; colors turn harsher and small objects appear in duplicate, which degrades FID yet inflates CLIPSIM as the model over-expresses caption tokens.
In the full model \textit{(\romannumeral 1)}, frequency scheduling suppresses glitches, AWP keeps the solution near a stable anchor, and both Procrustes and the blending loss guide the search along the natural-image manifold, trading a few points of raw caption similarity for substantially higher visual quality. The qualitative grid mirrors the numbers: \textit{(\romannumeral 1)} is the only variant that is simultaneously aesthetic, coherent, and semantically faithful.
\input{figs/ablation}

\Cref{fig:ablation} traces a prompt through 400 inversion steps to show how the most critical components influence the optimization. Two broad patterns emerge. \textit{(\romannumeral 1)} Frequency scheduling governs the refinement path: when it is present (top two rows) the image is generated in a coarse-to-fine order where color appears first, then shapes, then texture; without it (rows 3 \& 4) high-frequency stripes appear almost immediately and persist. \textit{(\romannumeral 2)} AWP mitigates cumulative drift: when it is present (rows 1 \& 3) the global scene layout stays stable throughout optimization, whereas its absence (rows 2 and 4) lets distortions and noise grow with every iteration.
The variant lacking both safeguards shows the combined failure modes, underscoring their complementary roles.

%% file: tables/main-results.tex
\definecolor{headerbg}{gray}{0.97}
\begin{table}
  \centering
  \renewcommand\arraystretch{1.2}
    \caption{\small{\textbf{MS-COCO text-to-image generation results}. FID (lower is better), CLIPSIM and IS (both higher are better), along with model sizes.}}     \label{tab:main_results}
  \resizebox{\textwidth}{!}{%
    \begin{tabular}{
      l 
      ccc 
      ccc 
      >{\columncolor{headerbg}}c
      >{\columncolor{headerbg}}c 
      >{\columncolor{headerbg}}c 
      >{\columncolor{headerbg}}c 
      >{\columncolor{headerbg}}c
    }
      \toprule
                          & \cite{betker2023improving} & \cite{nichol2022glide} & \cite{rombach2022high}   & \cite{ganz2023clipaggeneratorfreetexttoimagegeneration}  & \cite{ganz2024texttoimagegenerationenergybasedclip} & \cite{ganz2024texttoimagegenerationenergybasedclip} & \cite{kazemi2024learn} & \cite{fort2025direct} & \cite{fort2025direct} & \\
                          & DALL-E                     & GLIDE                  & LDM-KL-8                 & CLIPAG                & EB-CLIP                          & EB-CLIP                & CLIP-Inv                   & DAS                   & DAS                   & \tbf{CLIP$^{-1}$} \\
                          &                            &                        &                          & ViT                   & ViT                              & XXL                    & ViT                        & ViT                   & Ensemble              &                      \\
      \cmidrule(lr){2-4} \cmidrule(lr){5-7} \cmidrule(lr){8-11}
      Decoder-free        & \negxmark                  & \negxmark              & \negxmark                & \poscheckmark         & \poscheckmark                    & \poscheckmark          & \poscheckmark              & \poscheckmark         & \poscheckmark         & \poscheckmark        \\
      Tuning-free         & \negxmark                  & \negxmark              & \negxmark                & \negxmark             & \negxmark                        & \negxmark              & \poscheckmark              & \poscheckmark         & \poscheckmark         & \poscheckmark        \\ 
      \# Train params (M) & 12000                      & 6000                   & 1450                     & 88                    & 88                               & 1200                    & 0                          & 0                     & 0                     & \textbf{0}          \\ 
      \# Tot params (M)   & 12000                      & 6000                   & 1450                     & 150                   & 150                              & 846                    & 150                        & 150                   & 3x150                 & \textbf{150}          \\
      \cdashlinelr{1-12}
      FID$(\downarrow)$   & 27.5                       & 12.2                   & 23.3                     & 42.3                  & 68.3                             & 23.4                   & 140.1                          & 161.8                 & 121.6                 & \textbf{72.5}                \\
      CLIPSIM$(\uparrow)$ & --                         & --                     & --                       & 34.7                  & 34.5                             & 33.5                   & \textbf{61.4}                          & 22.7                  & 36.9                  & 38.6        \\
      IS$(\uparrow)$      & 17.9                       & --                     & 20.03                    & 18.7                    & --                               & --                     & 4.8                          & 5.7                   & 8.36                  & \textbf{9.5}         \\
      \bottomrule                                                                                                                                                                                                                                                                                              \\
    \end{tabular}
  }
\end{table}

%% file: figs/ablation.tex
\begin{figure}[b]
%\vspace{-20pt}
\centering
\begin{subfigure}[t]{\textwidth}
\centering
\begin{tabular}{@{}c@{\hspace{1pt}}c@{\,}c@{\,}c@{\,}c@{\,}c@{\,}c@{\,}c@{\,}c@{}}
    &\scriptsize \parbox[c]{0.1055\linewidth}{\centering \texttt{$i=0$}} & 
    \scriptsize \parbox[c]{0.1055\linewidth}{\centering \texttt{$i=50$}} & 
    \scriptsize \parbox[c]{0.1055\linewidth}{\centering \texttt{$i=100$}} & 
    \scriptsize \parbox[c]{0.1055\linewidth}{\centering \texttt{$i=150$}} &        
    \scriptsize \parbox[c]{0.1055\linewidth}{\centering \texttt{$i=200$}} & 
    \scriptsize \parbox[c]{0.1055\linewidth}{\centering \texttt{$i=250$}} & 
    \scriptsize \parbox[c]{0.1055\linewidth}{\centering \texttt{$i=300$}} & 
    \scriptsize \parbox[c]{0.1055\linewidth}{\centering \texttt{$i=400$}} \\
    \vspace{-2pt}
    \rotatebox{90}{\parbox[c]{0.1055\linewidth}{\centering \scriptsize \tbf{CLIP$^{-1}$}}}&
    \includegraphics[width=0.1055\linewidth]{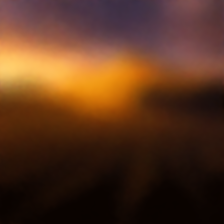} & 
    \includegraphics[width=0.1055\linewidth]{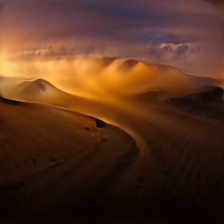} & 
    \includegraphics[width=0.1055\linewidth]{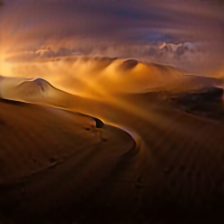} & 
    \includegraphics[width=0.1055\linewidth]{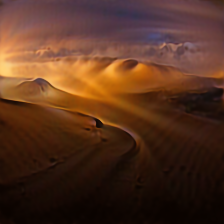} & 
    \includegraphics[width=0.1055\linewidth]{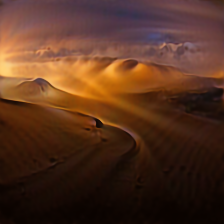} & 
    \includegraphics[width=0.1055\linewidth]{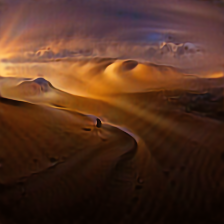} & 
    \includegraphics[width=0.1055\linewidth]{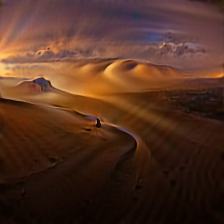} & 
    \includegraphics[width=0.1055\linewidth]{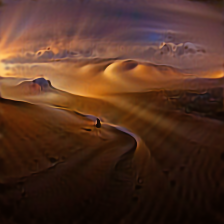} \\
    \vspace{-2pt}
    \rotatebox{90}{\parbox[c]{0.1055\linewidth}{\centering \scriptsize w/o AWP}}&
    \includegraphics[width=0.1055\linewidth]{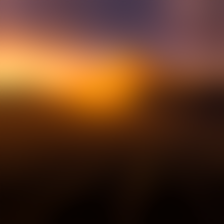} & 
    \includegraphics[width=0.1055\linewidth]{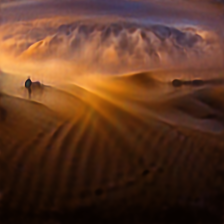} & 
    \includegraphics[width=0.1055\linewidth]{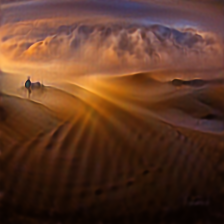} & 
    \includegraphics[width=0.1055\linewidth]{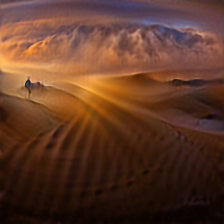} & 
    \includegraphics[width=0.1055\linewidth]{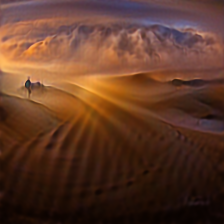} & 
    \includegraphics[width=0.1055\linewidth]{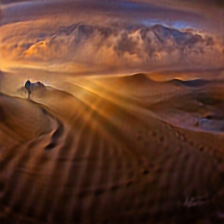} & 
    \includegraphics[width=0.1055\linewidth]{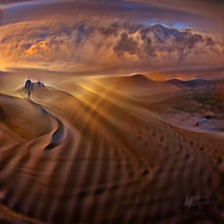} & 
    \includegraphics[width=0.1055\linewidth]{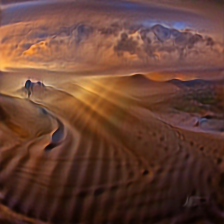} \\
    \vspace{-2pt}
    \rotatebox{90}{\parbox[c]{0.1055\linewidth}{\centering \scriptsize w/o Freq. Opt}} &

   \includegraphics[width=0.1055\linewidth]{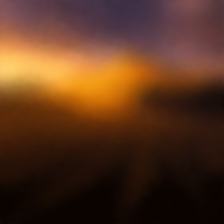} & 
    \includegraphics[width=0.1055\linewidth]{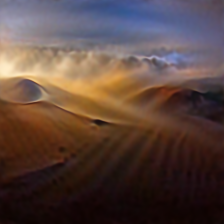} & 
    \includegraphics[width=0.1055\linewidth]{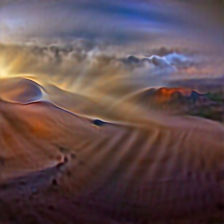} & 
    \includegraphics[width=0.1055\linewidth]{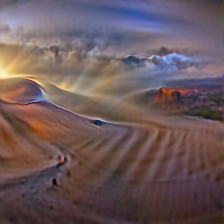} & 
    \includegraphics[width=0.1055\linewidth]{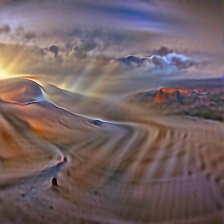} & 
    \includegraphics[width=0.1055\linewidth]{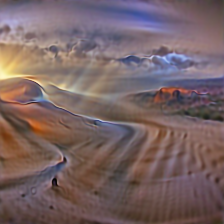} & 
    \includegraphics[width=0.1055\linewidth]{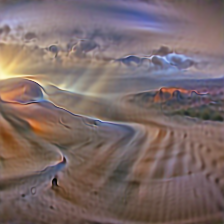} & 
    \includegraphics[width=0.1055\linewidth]{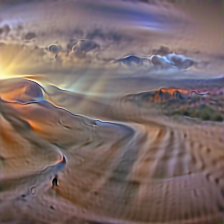} \\
    \vspace{-2pt}
    \rotatebox{90}{\parbox[c]{0.1055\linewidth}{\centering \scriptsize w/o both}} &
    
   \includegraphics[width=0.1055\linewidth]{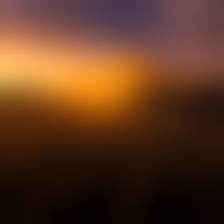} & 
    \includegraphics[width=0.1055\linewidth]{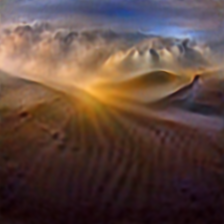} & 
    \includegraphics[width=0.1055\linewidth]{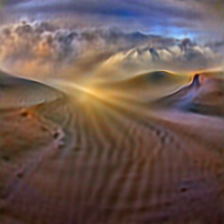} & 
    \includegraphics[width=0.1055\linewidth]{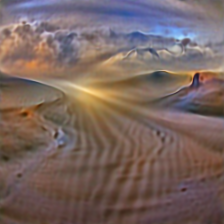} & 
    \includegraphics[width=0.1055\linewidth]{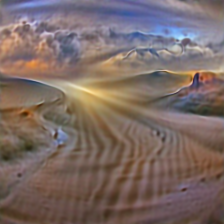} & 
    \includegraphics[width=0.1055\linewidth]{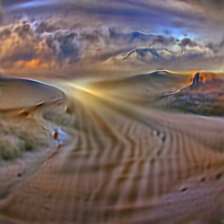} & 
    \includegraphics[width=0.1055\linewidth]{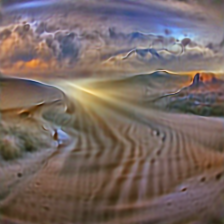} & 
    \includegraphics[width=0.1055\linewidth]{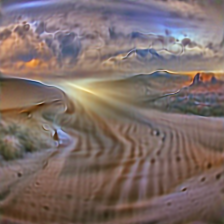} \\
  \end{tabular}
  \label{fig:your_label}
  % \subcaption*{\centering\scriptsize{ \texttt{\guillemotleft Panoramic view of a vast desert landscape at sunset with dramatic lighting, detailed dunes and a wide cinematic composition\guillemotright}}}
  \subcaption*{\centering\scriptsize{ \texttt{\guillemotleft Panoramic view of a vast desert landscape at sunset with dramatic lighting and detailed dunes\guillemotright}}}
\end{subfigure}
\caption{\small{\textbf{Qualitative ablation.}} Text-to-image synthesis of a desert landscape over $400$ iterations, comparing our full method (top row) with ablations: without AWP, without frequency-based optimization, and without both. }
\label{fig:ablation}
\end{figure}

%% file: sec/04_related_work.tex
\section{Related work}
\label{sec:related}
\minisection{Image generative models} 
Modern image generators fall into three broad families: GANs, diffusion models, and normalizing flows.
GANs train a generator to fool a discriminator with realistic samples, a strategy refined from early DCGANs~\cite{radford2015unsupervised} to StyleGAN-T~\cite{sauer2023stylegan} and BigGAN~\cite{brocklarge}. Diffusion models begin with pure noise and iteratively denoise it back to an image; latent diffusion~\cite{rombach2022high}, GLIDE~\cite{nichol2022glide}, DALL-E 2~
~\cite{ramesh2022hierarchical} and DALL-E 3~\cite{betker2023improving} differ mainly in how they compress the signal and guide it with text.
Normalizing flows (e.g., \cite{kingma2018glow, esser2024scaling}) learn a chain of invertible transforms so that sampling the base Gaussian and running the reverse pass yields data in a few steps.
Despite these distinct mechanics, every approach relies on a latent-to-image decoder: the generator in GANs, the denoising network in diffusion, or the reverse flow in normalizing-flow models. Our work removes this requirement altogether by directly inverting a frozen discriminative encoder (CLIP).

\minisection{Model inversion} 
Several recent works attempt to repurpose CLIP for image generation by inverting its embeddings. The earliest approach~\cite{kazemi2024learn} directly optimises randomly initialised pixels to minimise the cosine distance between the CLIP embedding of the image and that of a target text prompt. GALIP~\cite{tao2023galipgenerativeadversarialclips} introduces a CLIP-conditioned GAN framework, training both a generator and discriminator to enable fast, controllable synthesis with fewer parameters and less data than large-scale diffusion models. CLIPAG~\cite{ganz2023clipaggeneratorfreetexttoimagegeneration} follows a decoder-free design but applies adversarial fine-tuning to CLIP itself to improve generation quality. This idea is further extended in EB-CLIP~\cite{ganz2024texttoimagegenerationenergybasedclip}, where the generation is formulated as energy minimisation in CLIP's joint image-text space. Concurrently with our work, DAS~\cite{fort2025direct} shows that a frozen CLIP can be inverted without training by optimising at multiple spatial resolutions. Like us, DAS reveals generative priors within discriminative models, but differs by operating directly in pixel space rather than through a frequency-aware implicit representation.

\minisection{Adversarial robustness} Adversarial training has become a core strategy for improving model robustness, with~\cite{madry2017towards} introducing the first widely adopted method using input perturbations. TRADES~\cite{zhang2019theoretically} extended this by leveraging KL divergence to balance accuracy and robustness, later refined by~\cite{cui2023decoupled} to address its asymmetry. Beyond input-space attacks, AWP~\cite{wu2020adversarial} proposed perturbing model weights during training, improving generalisation by flattening the loss landscape. While primarily used for classification, recent work~\cite{mirza2024shedding} suggests robust models also encode stronger generative priors.

\minisection{Image modeling with INRs}
Implicit Neural Representations model images as continuous functions that map spatial coordinates $(i, j)$ to RGB values via a neural network, typically an MLP. To capture fine detail, they rely on frequency-aware components such as positional encodings~\cite{tancik2020fourierfeaturesletnetworks} or periodic activation functions like SIREN~\cite{sitzmann2020implicit}. However, fixed-frequency activations limit adaptability, motivating FINER~\cite{liu2024finer}, which introduces variable-periodic activations that dynamically adjust to local frequency content. We adopt FINER for its efficient representation, well-suited to our inversion task, and to mitigate the spectral bias that hampers high-fidelity reconstruction in standard INRs.

%% file: sec/06_conclusions.tex
\section{Conclusion}
This work presents CLIP$^{-1}$, a decoder-free and training-free approach that inverts the CLIP image encoder for text-to-image synthesis using implicit neural representations (INRs). Rather than requiring a generative decoder, we show that CLIP, when combined with a frequency-aware INR and a lightweight alignment mechanism, can guide image synthesis directly from text prompts.
While our results are not intended to compete with state-of-the-art generative models in absolute quality metrics, they highlight an underexplored and surprising capability: a frozen discriminative model like CLIP can be coaxed into producing coherent, semantically aligned images with no additional training or generative backbone. 
Moreover, the same framework supports zero-shot applications such as image reconstruction, controlled edits, and neural style transfer--all within a single unified setup. These findings open new directions for repurposing pretrained models for generation and suggest broader implications for model interpretability and robustness.
That said, the current approach relies entirely on the INR and associated losses to stay close to the natural image manifold; no explicit mechanism enforces this constraint. Incorporating a projection step back onto the manifold could improve fidelity and regularisation, and presents a promising direction for future work--especially toward understanding and interpreting the structure of CLIP’s embedding space.

\newpage

\minisection{Acknowledgment} This work was supported by projects PNRR MUR PE0000013-FAIR under the MUR National Recovery and Resilience Plan funded by the European Union - NextGenerationEU and PRIN 2022 project 20227YET9B ``AdVVent'' CUP code B53D23012830006. It was also partially supported by Sapienza research projects ``Prebunking'', ``Adagio'', ``Risk and Resilience factors in disadvantaged young people: a multi-method study in ecological and virtual environments'', ``BEAT (Better dEep leArning securiTy)'' and Seed of ERC grant ``MINT.AI''. 
Computing was supported by CINECA cluster under project Ge-Di HP10CRPUVC, RDM HP10C7YYL2 and the Sapienza Computer Science Department cluster. The authors would like to thank Andrea Salinetti for his initial work and bachelor's thesis on CLIP inversion.

%% file: sec/A_supmats.tex
\raggedbottom

\section{Supplementary Material}
This supplementary document expands on key aspects of our work by providing additional technical details and extended qualitative results. It is organized into six sections, each addressing a specific area that complements the main paper. \cref{sec:supp-awp} offers a detailed explanation of the adversarial weight perturbation (AWP) training procedure, including how the perturbation is computed and integrated into the overall training pipeline. \cref{sec:fullalgo} outlines the complete algorithm used for the text-to-image task. \cref{sec:supp-impl} describes the implementation setup, including model configurations, training parameters, and data preprocessing choices. \cref{sec:mscoco} presents the best and worst 500 generations from MS-COCO prompts, the same samples used in the paper's evaluation. \cref{sec:supp-qual} provides qualitative examples generated with different pre-trained ViT-B/32 models, illustrating the variability introduced by different backbone initializations. Finally, \cref{sec:supp-ablation} includes additional qualitative results from the ablation study, offering visual comparisons that highlight the contributions of individual components. This supplementary material is intended to support reproducibility and provide a deeper insight into our methodology and experimental findings.

\subsection{Further details on the AWP algorithm}
\label{sec:supp-awp}
We provide a detailed breakdown of the Adversarial Weight Perturbation (AWP) procedure and its integration into the INR training loop. \cref{alg:awp} outlines the core AWP mechanism: given the weights of the INR model $\phi$ and the input coordinates $\coords$ and a temporary clone $\widehat{\phi}$ is optimized to maximize the negative structural similarity index (SSIM) loss between the predicted output and a blurred version of the ground-truth image $\img$. The resulting adversarial perturbation $\Delta \phi$ is computed, normalized, and applied to the original weights $\phi$ to obtain the perturbed weights $\phi_{\text{adv}}$.

  % \begin{small}
    % \vspace{0pt}
\begin{center}
\scalebox{0.95}{
\setlength{\algomargin}{1.5em}
  \begin{algorithm}[H]
    \caption{\small{Adversarial Weight Perturbation (\textsc{compute\_awp})}} \label{alg:awp}
    \small{
      \textbf{Inputs: } INR model weights $\phi$, input coordinates $\coords$, target image $\img$\;
      $\widehat{\phi} \leftarrow \text{clone}(\phi)$ \tcp*{proxy model initialization}
      $\Loss_{\text{awp}} \leftarrow -\Loss_{\text{SSIM}}\bigl(f_{\widehat{\phi}}(\coords), \text{blur}(\img)\bigr)$ \tcp*{maximize the loss}
      $\text{Optimize } \widehat{\phi} \text{ w.r.t. } \Loss_{\text{awp}}$\;
      $\Delta \phi \leftarrow \widehat{\phi} - \phi$ \tcp*{compute perturbation}
      ${\Delta \phi} \leftarrow \gamma \cdot \frac{\|\phi\|}{\|\Delta \phi\| + \epsilon} \cdot \Delta \phi$ \tcp*{normalize and scale perturbation}

      \textbf{Return: } $\Delta \phi$\;
    }
  \end{algorithm}
}
\end{center}

\cref{alg:training} illustrates the incorporation of AWP into INR training. At each iteration, adversarial perturbations are computed using Algorithm~1 and then applied to the network. The overall training loss is a weighted combination of mean squared error (MSE), SSIM, and $\ell_1$ loss. This adversarial training scheme improves the robustness and generalization of the INR by encouraging consistency under weight-level perturbations, which are the gradients received by inverting CLIP when generating.

\begin{center}
\scalebox{0.95}{
  \begin{algorithm}[H]
    \caption{\small{INR Training with AWP}} \label{alg:training}
    \small{
      \textbf{Inputs: } target image $\img$, initial weights $\phi_0$, input coordinates $\coords$\;
      \textbf{Hyperparameters: } learning rate $\eta$, perturbation scale $\gamma$, iterations $N$\;
      \For {$k=1,\ldots,N$}{
        $\Delta \phi \leftarrow \textsc{compute\_awp}(\phi_{k}, \coords, \img)$\;
        $\phi_{\text{adv}} \leftarrow \phi_{k} + {\Delta \phi}$ \tcp*{apply perturbation}
        
        $f_{\phi_{\text{adv}}} \leftarrow$ model with weights $\phi_{\text{adv}}$\;
        $\hat{\img} \leftarrow f_{\phi_{\text{adv}}}(\coords)$\;
        
        $\Loss \leftarrow \alpha_1 \Loss_\text{MSE}(\hat{\img}, \text{blur}(\img)) + \alpha_2 \Loss_\text{SSIM}(\hat{\img}, \text{blur}(\img)) + \alpha_3 \Loss_{\text{L1}}(\hat{\img} - \text{blur}(\img))$\;
        \text{Update } $\phi_{\text{adv}}$ \text{ via optimizer step minimizing } $\Loss$\;
        
        $\phi_{k+1} \leftarrow \phi_{\text{adv}} - {\Delta \phi}$ \tcp*{restore original weights for next iteration}
      }
      \textbf{Return: } trained weights $\phi_N$\;
    }
  \end{algorithm}
}
\end{center}

% \end{small}

\subsection{Text-to-Image full pipeline}
\label{sec:fullalgo}
We optimize an implicit neural representation (INR) to synthesize an image that aligns with a given text prompt using CLIP. The procedure includes text and image retrieval, feature alignment, and iterative gradient-based optimization. Procrustes alignment and natural image constraints are enabled. \cref{alg:fullinversion} shows the detailed pipeline.
    \begin{figure}[H]
    \resizebox{0.90\linewidth}{!}{
    \setlength{\algomargin}{1.5em}
\begin{algorithm}[H]
\caption{Text-to-Image Synthesis}
\label{alg:fullinversion}
\textbf{Input:} Text prompt $\prompt$, \\
\textbf{Output:} Synthesized image ${\img}=f_{\phi^N}(\coords)$ \\
\;
 \tcc{Input pre-processing and alignment}
 Encode the input prompt $\embtext = \clipt{\prompt}$ \;
 % Encode auxiliary positive texts  to obtain $f_{\text{pos}}, f_{\text{neg}}$ \;
% \;
%  \tcc{Procrustes Alignment:}
Select top-$k$ matches to $\embtext$ in $\mathcal{D}$ to build $\mbf{E}_T, \mbf{E}_I \in \mathbb{R}^{d \times k}$\;
    Compute on the fly the orthogonal Procrustes rotation matrix   $R =  \min_{\mbf{R}} ||\mbf{R}\mbf{E}_T-\mbf{E}_I||_F \qquad \text{s.t.}\  \mbf{R}^\top\mbf{R}=\mbf{I}\ $\;
 Project $\embtext$ to visual domain: $\embtexttoim = \mbf{R}\clipt{\prompt}$\;
\;
 \tcc{Retrieve Initialization:}
 From dataset $\mathcal{D}$, retrieve image embedding $\clipi{\hat{\img}}$ of $\hat{\img}$ with caption closest to $\embtexttoim$\;
 Initialize INR weights $\phi_0 \leftarrow \small{\textsc{INIT\_INR\_AWP}}(\hat{\img})$---see \cref{alg:training}\;
\;
 \tcc{Natural Image Constraints:}
 Retrieve top-$k$ natural images $\{\mbf{x}^\star_{j}\}$ near $\embtext$ in CLIP space\;
 Encode them to features $\{\mbf{e}^\star_{img,j}\}$\;
 Store the similarity to the input prompt $w_j = \textsc{CLIPSIM}(\mbf{x}^\star_{j}, \embtext)$\;
 Compute weighted average: $\mbf{e}^\star_{img} = \sum_j w_j\mbf{e}^\star_{img,j}$ where $w_j$ are normalized w/ softmax \;
\;
 \tcc{Optimizer Setup:}
 Initialize layer-wise optimizers with Gaussian learning rates (peak is $\gamma$) over INR depth.\;
 Schedule shifting of Gaussian center every $k$ steps\;
\;
\For{$i = 1$ to $T$}{
    \If{learning rate schedule triggers}
       {  Shift Gaussian center layer\;
    }
     Encode via CLIP the augmentations of the rendered INR:
     $$\quad \mbf{e}_i^\star = \frac{1}{n} \sum_{k=1}^{n} \clipi{\text{augment}(f_{\phi^k})}$$\;
     Compute total loss:$
        \Loss(\mbf{e}_i^\star, \embtexttoim) + \beta \Loss_{blend}\left( \mbf{e}_i^\star,\mbf{e}^\star_{img}\right)
    $\;
     Update $\phi$: $$\quad \phi_{n} = \phi_{n-1} -  \nabla_{\phi} \Big[\Loss(\mbf{e}_i^\star, \embtexttoim) + \beta \Loss_{blend}\left( \mbf{e}_i^\star,\mbf{e}^\star_{img}\right) \Big] $$
}
 \textbf{Return:} Final image ${\img}=f_{\phi^N}(\coords)$
\end{algorithm}}
\end{figure}
\subsection{Implementation details}
\label{sec:supp-impl}
\subsubsection{INR parameters and initialization}
We initialize our implicit neural representations (INRs) with \texttt{in\_features} = 2 and \texttt{out\_features} = 3, using five hidden layers of 256 units each. Sinusoidal parameterization is applied with \texttt{first\_omega} = 25 and \texttt{hidden\_omega} = 25 to enable high-frequency signal modeling. Training is performed using the Adam optimizer with a learning rate of $1 \times 10^{-4}$, and a cosine annealing schedule via \texttt{torch.optim.lr\_scheduler.CosineAnnealingWarmRestarts} with a restart period of 100 iterations.

We apply Adversarial Weight Perturbation (AWP) using a proxy optimizer with the same learning rate and a perturbation strength of $\alpha = 0.01$. The ground-truth image $\img$ is preprocessed using a Gaussian blur with \texttt{kernel\_size} = 101 and $\sigma \in (10.0, 20.0)$ to provide a smoother supervision signal.

The training loss combines mean squared error (MSE), structural similarity (SSIM), and $\ell_1$ reconstruction loss, weighted, respectively, by $\alpha_1 = 0.85$, $\alpha_2 = 0.25$, and $\alpha_3 = 0.25$.

\subsubsection{Text-to-image inversion parameters}
Our method is built upon a \textit{ViT-B/32} backbone initialized with the default OpenAI weights. We perform 400 inversion steps using the \textit{AdamW} optimizer (without \textit{AMSGrad}) and a learning rate of $2 \times 10^{-4}$. During INR optimization, we employ a Gaussian scheduling strategy focused on layers $[0, 1, 2]$, with gradient norm clipping thresholds set to $[1.0, 0.5, 0.2]$ respectively. This schedule is refreshed every 70 iterations to preserve both stability and optimization efficiency over time.

The loss function incorporates hyperparameters $\beta = 0.5$ and $k = 8$, balancing the trade-offs between reconstruction fidelity and robustness.
To promote generalization, we apply data augmentation by generating 32 variations per input sample.
For spatial alignment, we use Orthogonal Procrustes analysis over the nearest $p = 256$ elements.
Following Stable Diffusion~\cite{jagielskimeasuring}, we guide the CLIP inversion process by appending auxiliary textual prompts that explicitly describe desired image characteristics. This strategy improves the fidelity and perceptual quality of the generated outputs.

The complete text-to-image synthesis pipeline executes in approximately \textit{1 minute} and \textit{18 seconds} on a single NVIDIA RTX 4060. The entire process requires around \textit{3200MiB} of VRAM.

\clearpage
\subsection{Best vs. Worst MS-COCO generations}
\label{sec:mscoco}
\begin{figure}[H]
    \centering
    \includegraphics[width=1\linewidth]{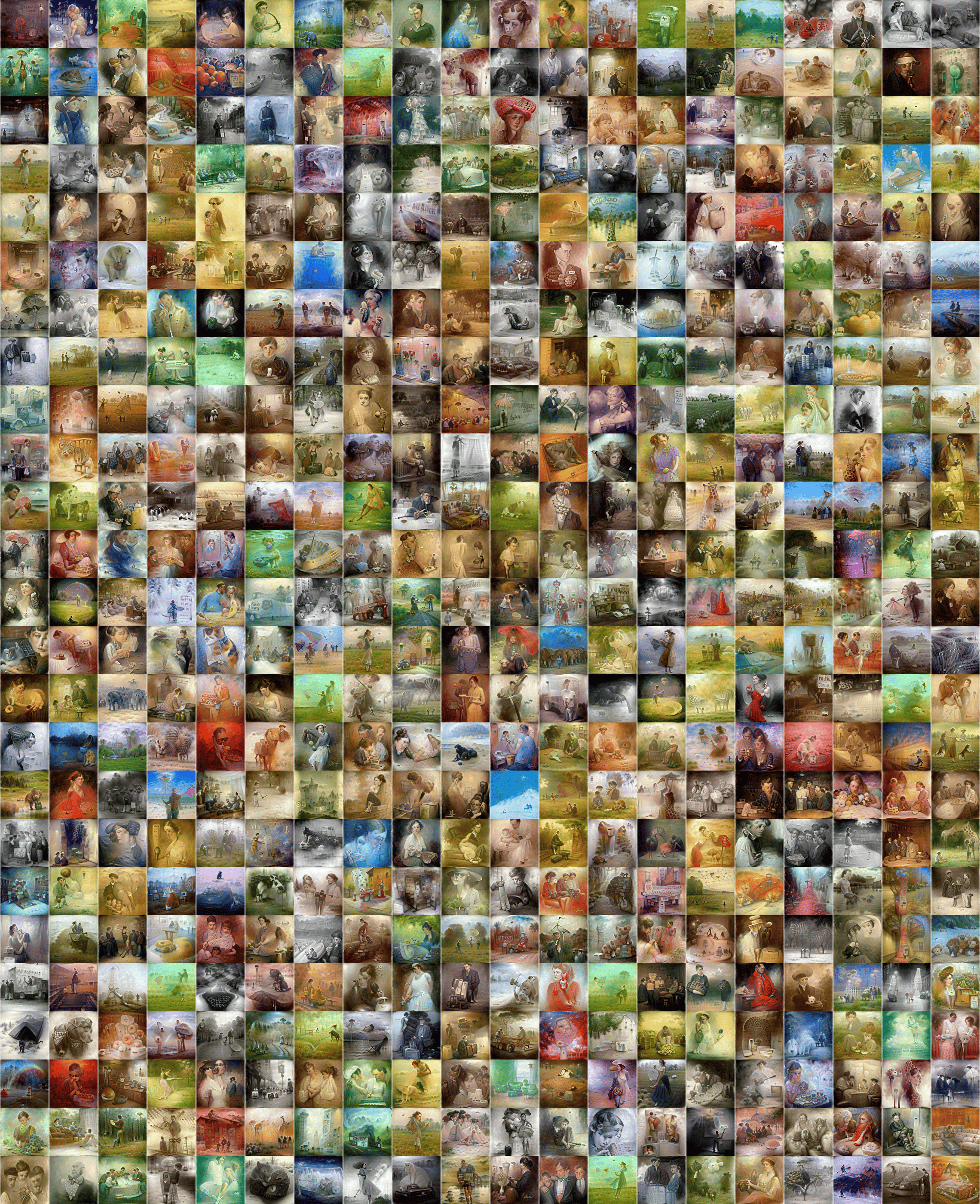}
    \caption{\small{\tbf{Best generations on MS-COCO prompts}. CLIP Score ranging from 45.5 to 53.9 (mean: 47.2)}}
    \label{fig:best}
\end{figure}

\begin{figure}[H]
    \centering
    \includegraphics[width=1\linewidth]{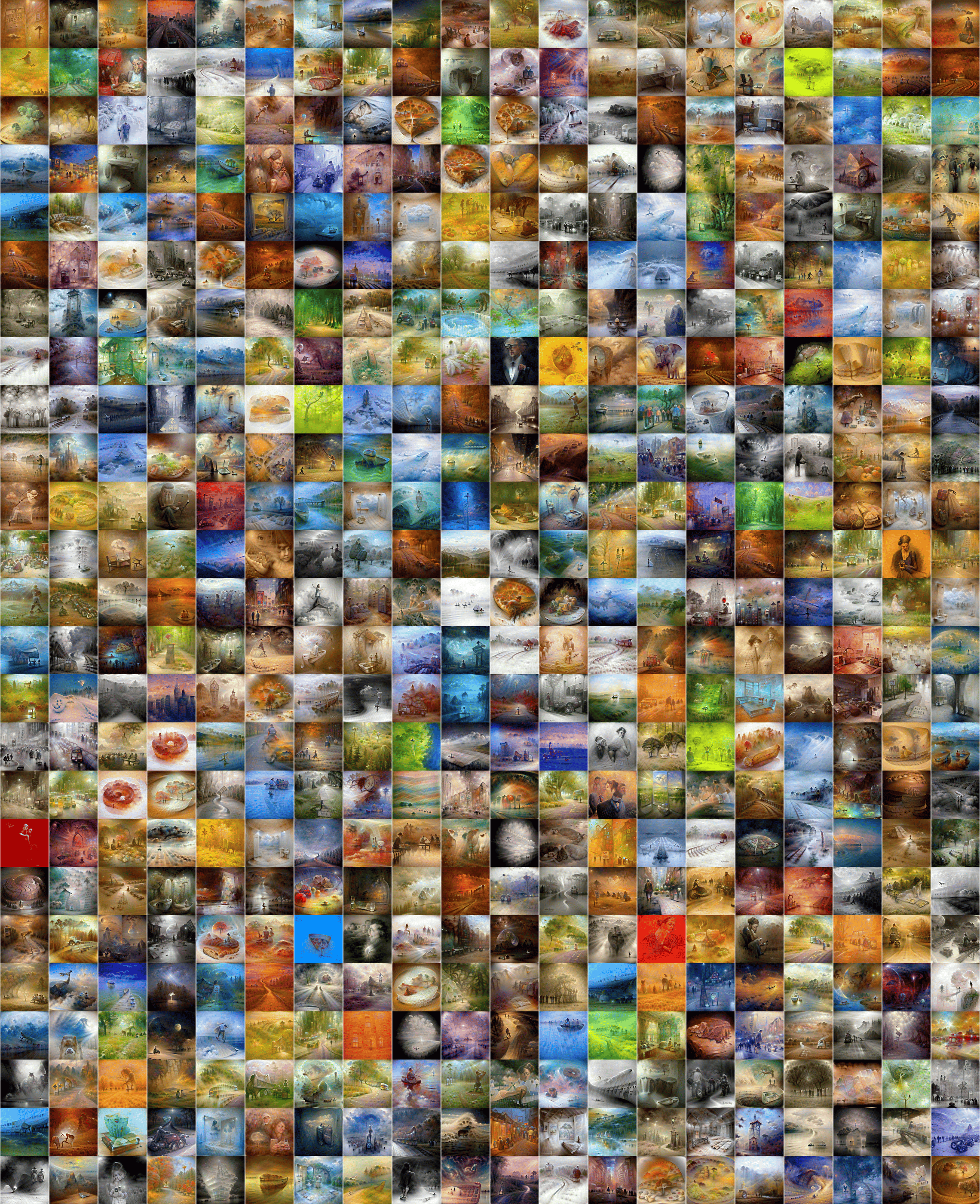}
    \caption{\small{\tbf{Worst generations on MS-COCO prompts}}. CLIP Score ranging from 23.0 to 32.2 (mean: 30.6)}
    \label{fig:worst}
\end{figure}

To qualitatively assess model's performance, this section presents the 500 highest and 500 lowest scoring generations based on CLIP Score, using prompts from the MS-COCO captions dataset. \cref{fig:best} shows the best generations that achieved CLIP scores between 45.5 and 53.9 (mean: 47.2). \cref{fig:worst} shows the worst, ranging from 23.0 to 32.2 (mean: 30.6). These examples illustrate the range of output quality, from strong semantic alignment to notable failure cases.

%We include an additional regularization term, $\Loss_{\text{const}}$, defined as the cosine distance between the CLIP embedding of the synthesized image, computed after applying the augmentations described in \cref{eq:emb-aug}, and that of the quality-related textual constraint.
\clearpage
\subsection{Qualitative samples under different CLIP models}

\label{sec:supp-qual}
\begin{figure}[H]
\begin{tabular}{@{\,}c@{\,}c@{\,}c@{\,}c@{\,}c@{\,}c@{\,}c@{\,}c@{\,}}
  \scriptsize \parbox[c]{0.12\linewidth}{\centering \texttt{CLIPAG \\ \cite{ganz2023clipaggeneratorfreetexttoimagegeneration}}} &
  \scriptsize \parbox[c]{0.12\linewidth}{\centering \texttt{CLIP-JEM \\\cite{ganz2024texttoimagegenerationenergybasedclip}}} &
  \scriptsize \parbox[c]{0.12\linewidth}{\centering \texttt{DAS \\ \cite{fort2025direct}}} &
  \scriptsize \parbox[c]{0.12\linewidth}{\centering \texttt{CLIP-Inv \cite{kazemi2024learn}}} &
  \scriptsize \parbox[c]{0.12\linewidth}{\centering \tbf{CLIP$^{-1}$} \\\texttt{(ViT-B/32})} &
  \scriptsize \parbox[c]{0.12\linewidth}{\centering \tbf{CLIP$^{-1}$} \\\texttt{(RESNET)}} &
  \scriptsize \parbox[c]{0.12\linewidth}{\centering \tbf{CLIP$^{-1}$} \\\texttt{w/\cite{ganz2023clipaggeneratorfreetexttoimagegeneration} }} &
  \scriptsize \parbox[c]{0.12\linewidth}{\centering \tbf{CLIP$^{-1}$} \\\texttt{w/\cite{ganz2024texttoimagegenerationenergybasedclip} (XXL)}} \\

  \includegraphics[width=0.12\linewidth]{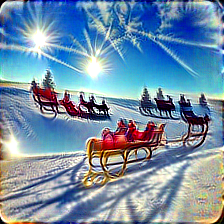} &
  \includegraphics[width=0.12\linewidth]{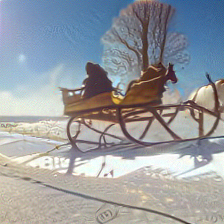} &
  \includegraphics[width=0.12\linewidth]{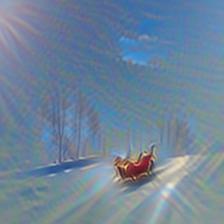} &
  \includegraphics[width=0.12\linewidth]{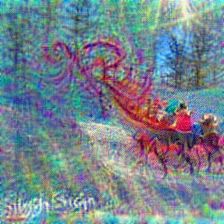} &
  \includegraphics[width=0.12\linewidth]{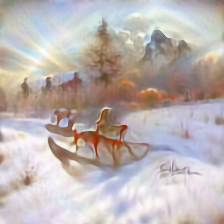} &
  \includegraphics[width=0.12\linewidth]{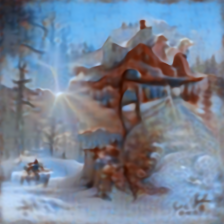} &
  \includegraphics[width=0.12\linewidth]{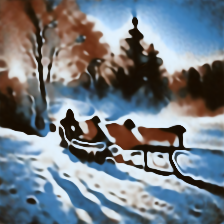} &
  \includegraphics[width=0.12\linewidth]{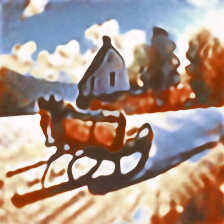} \\
  \multicolumn{8}{c}{\scriptsize \texttt{\guillemotleft Sleigh Ride On A Sunny Day\guillemotright}} \\

  \includegraphics[width=0.12\linewidth]{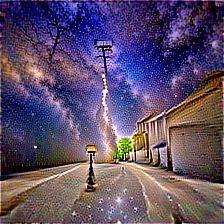} &
  \includegraphics[width=0.12\linewidth]{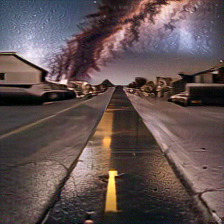} &
  \includegraphics[width=0.12\linewidth]{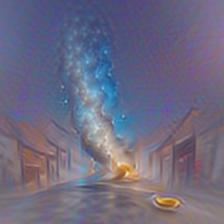} &
  \includegraphics[width=0.12\linewidth]{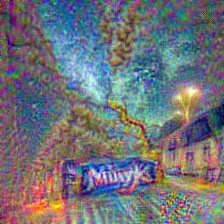} &
  \includegraphics[width=0.12\linewidth]{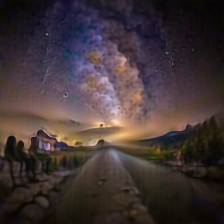} &
  \includegraphics[width=0.12\linewidth]{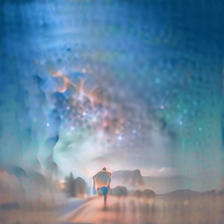} &
  \includegraphics[width=0.12\linewidth]{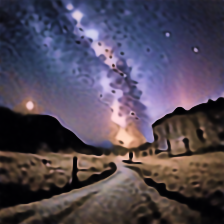} &
  \includegraphics[width=0.12\linewidth]{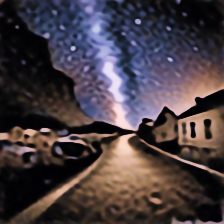} \\
  \multicolumn{8}{c}{\scriptsize \texttt{\guillemotleft Milky Way Over a Street\guillemotright}} \\

  \includegraphics[width=0.12\linewidth]{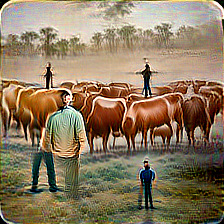} &
  \includegraphics[width=0.12\linewidth]{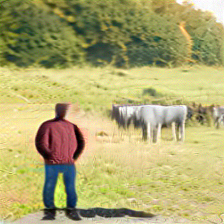} &
  \includegraphics[width=0.12\linewidth]{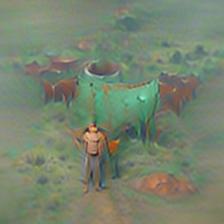} &
  \includegraphics[width=0.12\linewidth]{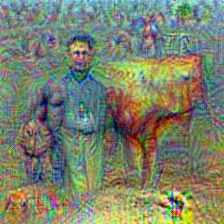} &
  \includegraphics[width=0.12\linewidth]{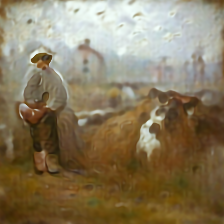} &
  \includegraphics[width=0.12\linewidth]{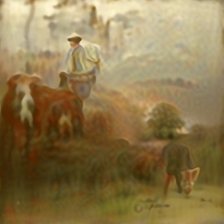} &
  \includegraphics[width=0.12\linewidth]{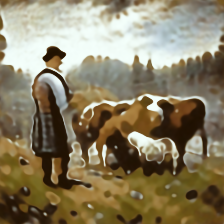} &
  \includegraphics[width=0.12\linewidth]{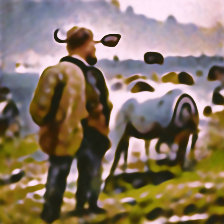} \\
  \multicolumn{8}{c}{\scriptsize \texttt{\guillemotleft Man standing near a cattle\guillemotright}} \\

  \includegraphics[width=0.12\linewidth]{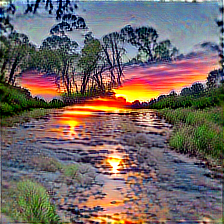} &
  \includegraphics[width=0.12\linewidth]{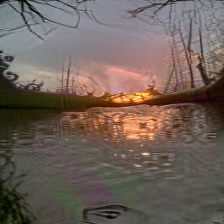} &
  \includegraphics[width=0.12\linewidth]{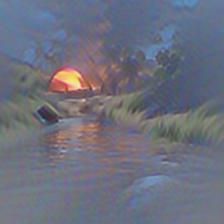} &
  \includegraphics[width=0.12\linewidth]{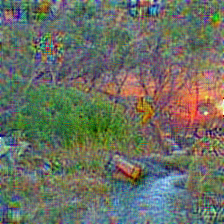} &
  \includegraphics[width=0.12\linewidth]{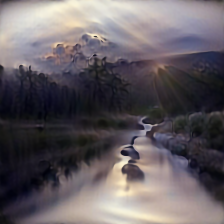} &
  \includegraphics[width=0.12\linewidth]{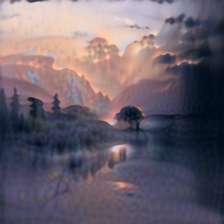} &
  \includegraphics[width=0.12\linewidth]{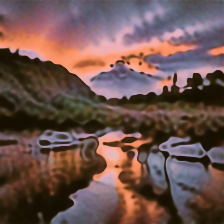} &
  \includegraphics[width=0.12\linewidth]{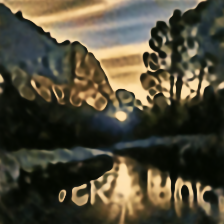} \\
  \multicolumn{8}{c}{\scriptsize \texttt{\guillemotleft Sunset seen from the creek\guillemotright}} \\

  \includegraphics[width=0.12\linewidth]{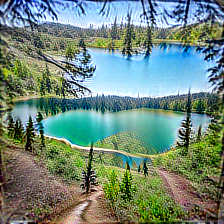} &
  \includegraphics[width=0.12\linewidth]{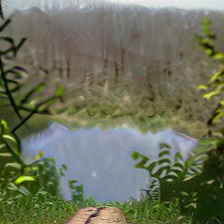} &
  \includegraphics[width=0.12\linewidth]{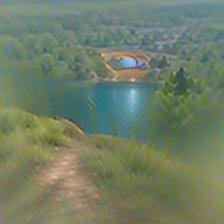} &
  \includegraphics[width=0.12\linewidth]{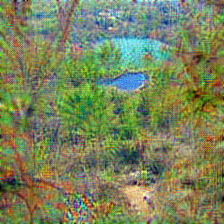} &
  \includegraphics[width=0.12\linewidth]{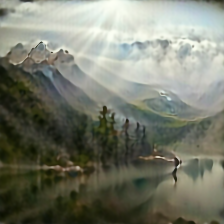} &
  \includegraphics[width=0.12\linewidth]{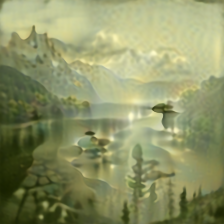} &
  \includegraphics[width=0.12\linewidth]{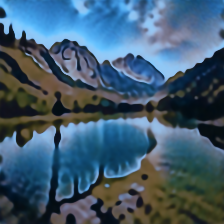} &
  \includegraphics[width=0.12\linewidth]{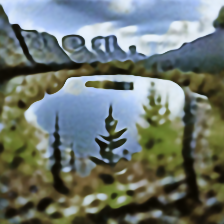} \\
  \multicolumn{8}{c}{\scriptsize \texttt{\guillemotleft View of a lake from a hiking trail\guillemotright}} \\

  \includegraphics[width=0.12\linewidth]{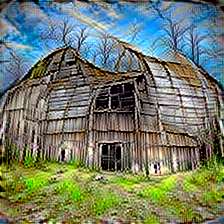} &
  \includegraphics[width=0.12\linewidth]{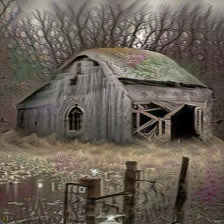} &
  \includegraphics[width=0.12\linewidth]{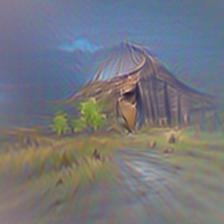} &
  \includegraphics[width=0.12\linewidth]{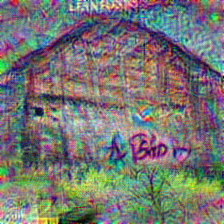} &
  \includegraphics[width=0.12\linewidth]{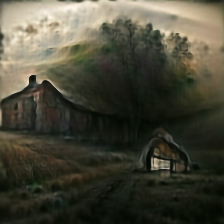} &
  \includegraphics[width=0.12\linewidth]{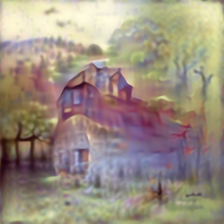} &
  \includegraphics[width=0.12\linewidth]{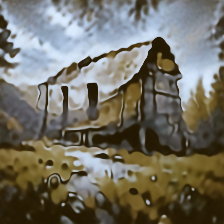} &
  \includegraphics[width=0.12\linewidth]{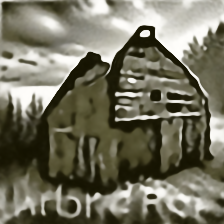} \\
  \multicolumn{8}{c}{\scriptsize \texttt{\guillemotleft Abandoned old barn\guillemotright}} \\

  \includegraphics[width=0.12\linewidth]{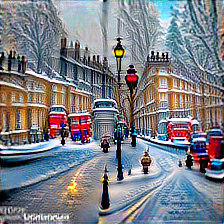} &
  \includegraphics[width=0.12\linewidth]{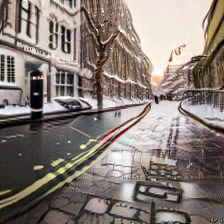} &
  \includegraphics[width=0.12\linewidth]{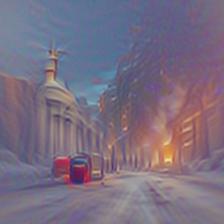} &
  \includegraphics[width=0.12\linewidth]{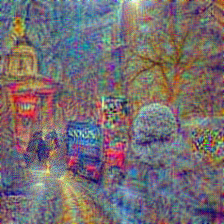} &
  \includegraphics[width=0.12\linewidth]{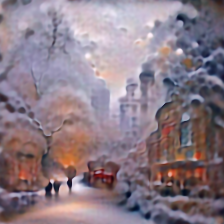} &
  \includegraphics[width=0.12\linewidth]{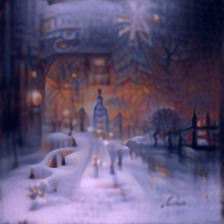} &
  \includegraphics[width=0.12\linewidth]{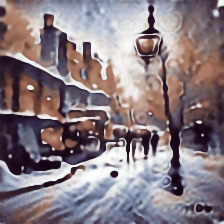} &
  \includegraphics[width=0.12\linewidth]{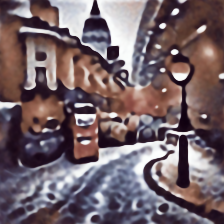} \\
  \multicolumn{8}{c}{\scriptsize \texttt{\guillemotleft Streets of London in winter\guillemotright}} \\

  \includegraphics[width=0.12\linewidth]{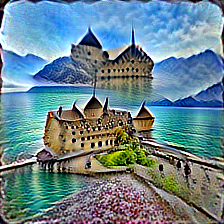} &
  \includegraphics[width=0.12\linewidth]{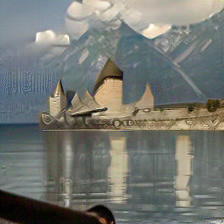} &
  \includegraphics[width=0.12\linewidth]{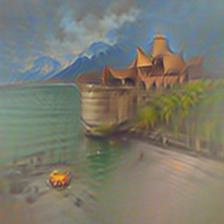} &
  \includegraphics[width=0.12\linewidth]{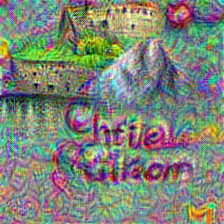} &
  \includegraphics[width=0.12\linewidth]{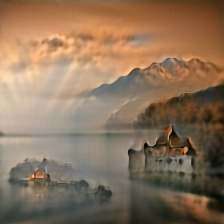} &
  \includegraphics[width=0.12\linewidth]{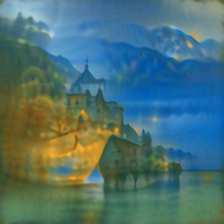} &
  \includegraphics[width=0.12\linewidth]{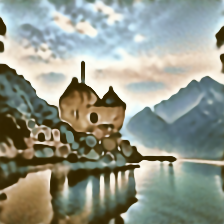} &
  \includegraphics[width=0.12\linewidth]{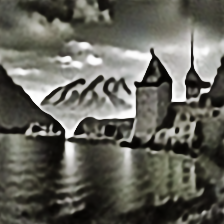} \\
  \multicolumn{8}{c}{\scriptsize \texttt{\guillemotleft View of the Chateau de Chillon\guillemotright}} \\
\end{tabular}
  \caption{\small{\textbf{Qualitative comparison} of additional samples extending Figure 4 in the main paper.}}
\end{figure}
\clearpage

\subsection{Qualitative samples of the ablation study} \label{sec:supp-ablation}
\begin{figure}[H]
  \centering

  % Prima colonna (sinistra)
  \begin{minipage}[t]{0.48\textwidth}
    \centering
    \begin{tabular}{@{\,}c@{\,}c@{\,}c@{\,}c@{\,}c@{\,}}
      \scriptsize \parbox[c]{0.18\linewidth}{\centering \tbf{CLIP$^{-1}$}} &
      \scriptsize \parbox[c]{0.18\linewidth}{\centering \tbf{w/o Freq. Opt}}  &
      \scriptsize \parbox[c]{0.18\linewidth}{\centering \tbf{w/o AWP}}  &
      \scriptsize \parbox[c]{0.18\linewidth}{\centering \tbf{w/o F.O. + Procrustes}}  &
      \scriptsize \parbox[c]{0.18\linewidth}{\centering \tbf{w/o F.O. + Blending}}      \\
      \includegraphics[width=0.18\linewidth]{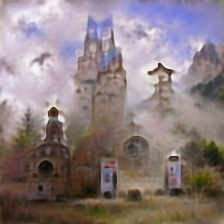} &
      \includegraphics[width=0.18\linewidth]{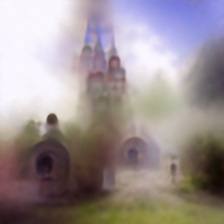} &
      \includegraphics[width=0.18\linewidth]{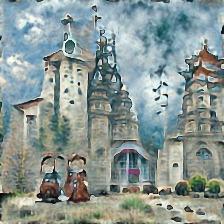} &
      \includegraphics[width=0.18\linewidth]{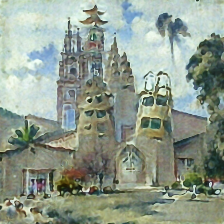} &          
      \includegraphics[width=0.18\linewidth]{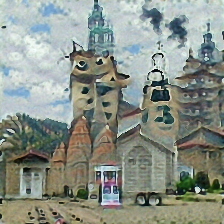} \\
      \includegraphics[width=0.18\linewidth]{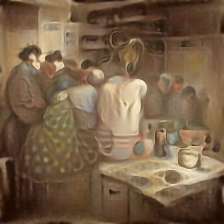} &
      \includegraphics[width=0.18\linewidth]{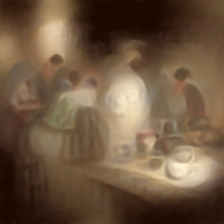} &
      \includegraphics[width=0.18\linewidth]{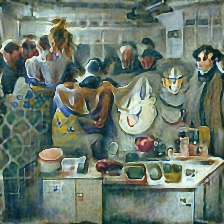} &
      \includegraphics[width=0.18\linewidth]{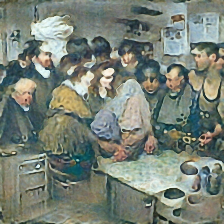} &          
      \includegraphics[width=0.18\linewidth]{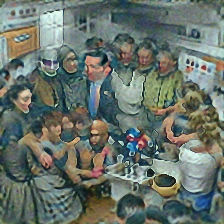} \\
      \includegraphics[width=0.18\linewidth]{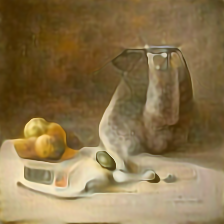} &
      \includegraphics[width=0.18\linewidth]{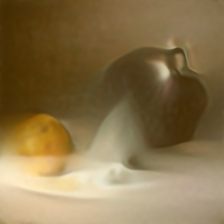} &
      \includegraphics[width=0.18\linewidth]{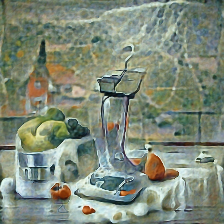} &
      \includegraphics[width=0.18\linewidth]{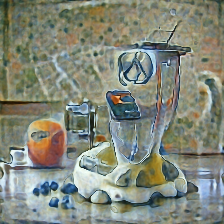} &          
      \includegraphics[width=0.18\linewidth]{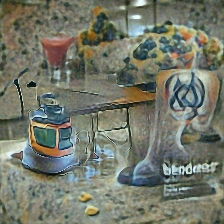} \\
      \includegraphics[width=0.18\linewidth]{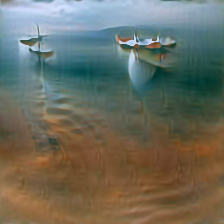} &
      \includegraphics[width=0.18\linewidth]{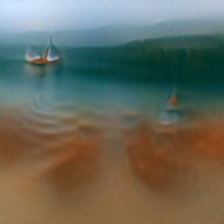} &
      \includegraphics[width=0.18\linewidth]{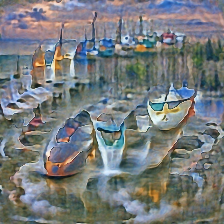} &
      \includegraphics[width=0.18\linewidth]{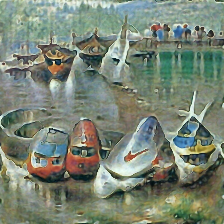} &          
      \includegraphics[width=0.18\linewidth]{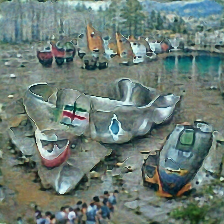} \\
            \includegraphics[width=0.18\linewidth]{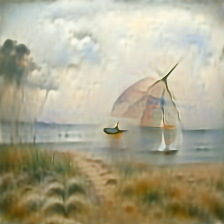} &
      \includegraphics[width=0.18\linewidth]{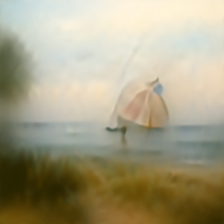} &
      \includegraphics[width=0.18\linewidth]{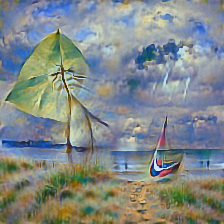} &
      \includegraphics[width=0.18\linewidth]{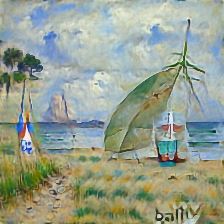} &          
      \includegraphics[width=0.18\linewidth]{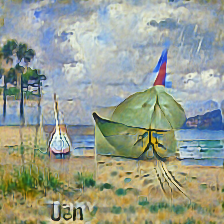} \\
            \includegraphics[width=0.18\linewidth]{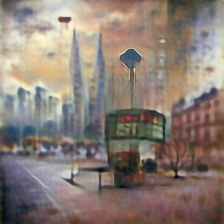} &
      \includegraphics[width=0.18\linewidth]{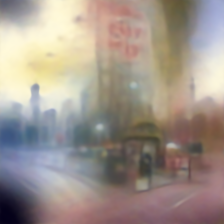} &
      \includegraphics[width=0.18\linewidth]{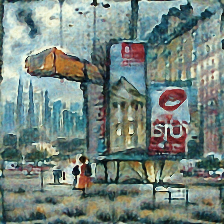} &
      \includegraphics[width=0.18\linewidth]{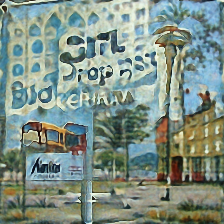} &          
      \includegraphics[width=0.18\linewidth]{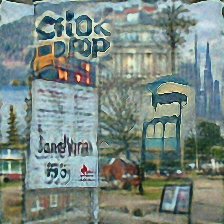} \\
            \includegraphics[width=0.18\linewidth]{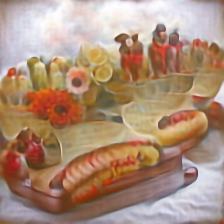} &
      \includegraphics[width=0.18\linewidth]{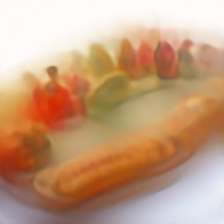} &
      \includegraphics[width=0.18\linewidth]{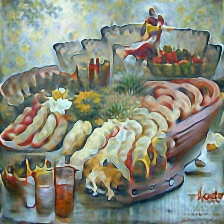} &
      \includegraphics[width=0.18\linewidth]{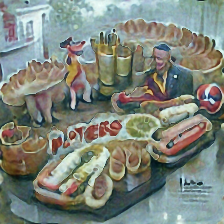} &          
      \includegraphics[width=0.18\linewidth]{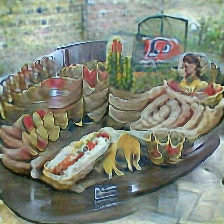} \\
            \includegraphics[width=0.18\linewidth]{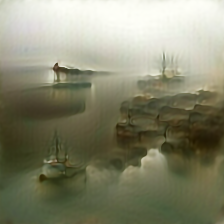} &
      \includegraphics[width=0.18\linewidth]{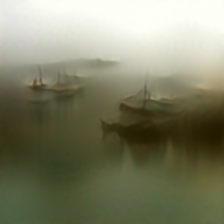} &
      \includegraphics[width=0.18\linewidth]{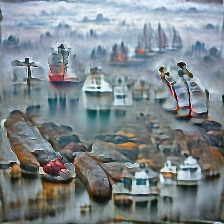} &
      \includegraphics[width=0.18\linewidth]{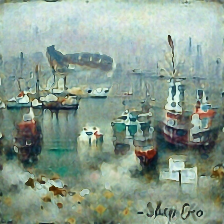} &          
      \includegraphics[width=0.18\linewidth]{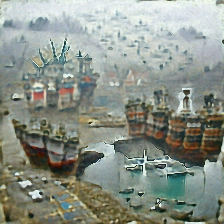} \\
            \includegraphics[width=0.18\linewidth]{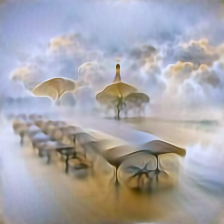} &
      \includegraphics[width=0.18\linewidth]{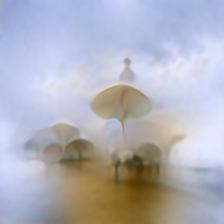} &
      \includegraphics[width=0.18\linewidth]{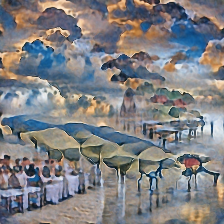} &
      \includegraphics[width=0.18\linewidth]{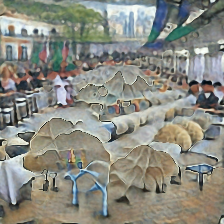} &          
      \includegraphics[width=0.18\linewidth]{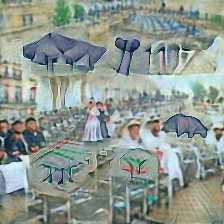} \\
            \includegraphics[width=0.18\linewidth]{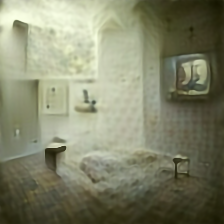} &
      \includegraphics[width=0.18\linewidth]{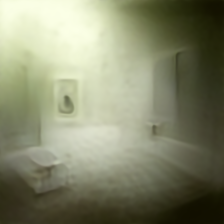} &
      \includegraphics[width=0.18\linewidth]{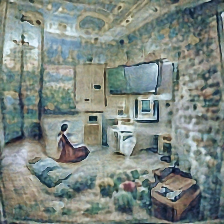} &
      \includegraphics[width=0.18\linewidth]{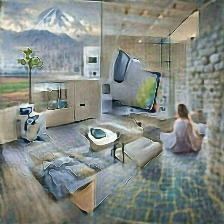} &          
      \includegraphics[width=0.18\linewidth]{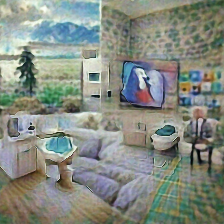} \\
    \end{tabular}
    \caption*{\small{\textbf{Left:} CLIP$^{-1}$ Ablation Rows 1-10}}
  \end{minipage}%
  \hfill
  % Seconda colonna (destra)
  \begin{minipage}[t]{0.48\textwidth}
    \centering
    \begin{tabular}{@{\,}c@{\,}c@{\,}c@{\,}c@{\,}c@{\,}}
      \scriptsize \parbox[c]{0.18\linewidth}{\centering \tbf{CLIP$^{-1}$}} &
      \scriptsize \parbox[c]{0.18\linewidth}{\centering \tbf{w/o Freq. Opt}}  &
      \scriptsize \parbox[c]{0.18\linewidth}{\centering \tbf{w/o AWP}}  &
      \scriptsize \parbox[c]{0.18\linewidth}{\centering \tbf{w/o F.O. + Procrustes}}  &
      \scriptsize \parbox[c]{0.18\linewidth}{\centering \tbf{w/o F.O. + Blending}}      \\
      \includegraphics[width=0.18\linewidth]{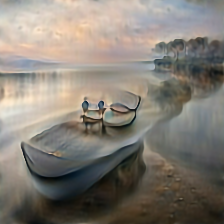} &
      \includegraphics[width=0.18\linewidth]{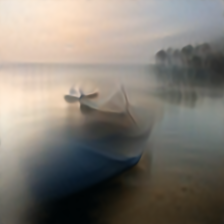} &
      \includegraphics[width=0.18\linewidth]{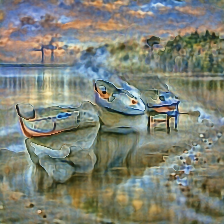} &
      \includegraphics[width=0.18\linewidth]{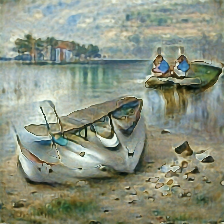} &          
      \includegraphics[width=0.18\linewidth]{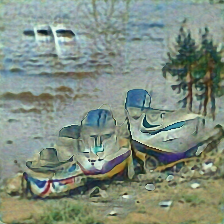} \\
      \includegraphics[width=0.18\linewidth]{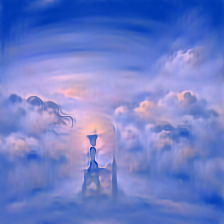} &
      \includegraphics[width=0.18\linewidth]{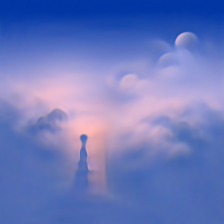} &
      \includegraphics[width=0.18\linewidth]{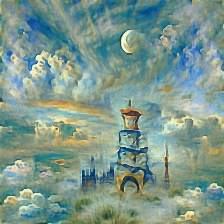} &
      \includegraphics[width=0.18\linewidth]{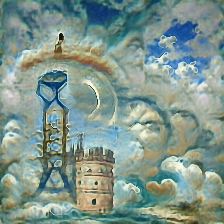} &          
      \includegraphics[width=0.18\linewidth]{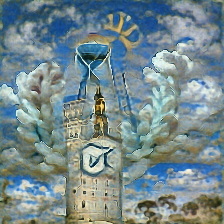} \\
      \includegraphics[width=0.18\linewidth]{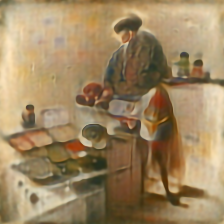} &
      \includegraphics[width=0.18\linewidth]{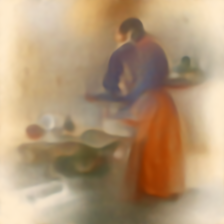} &
      \includegraphics[width=0.18\linewidth]{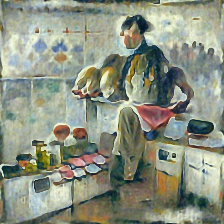} &
      \includegraphics[width=0.18\linewidth]{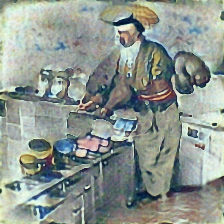} &          
      \includegraphics[width=0.18\linewidth]{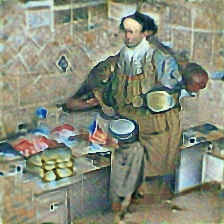} \\
      \includegraphics[width=0.18\linewidth]{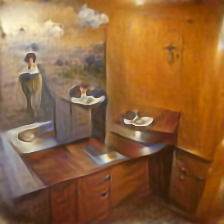} &
      \includegraphics[width=0.18\linewidth]{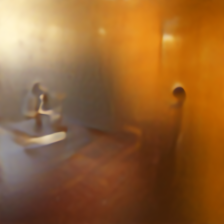} &
      \includegraphics[width=0.18\linewidth]{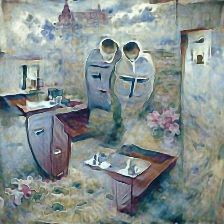} &
      \includegraphics[width=0.18\linewidth]{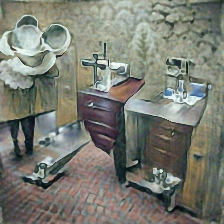} &          
      \includegraphics[width=0.18\linewidth]{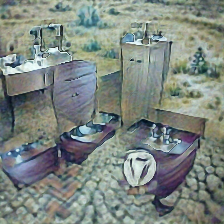} \\
            \includegraphics[width=0.18\linewidth]{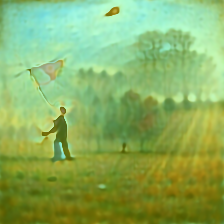} &
      \includegraphics[width=0.18\linewidth]{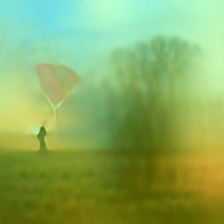} &
      \includegraphics[width=0.18\linewidth]{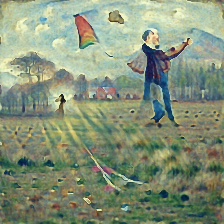} &
      \includegraphics[width=0.18\linewidth]{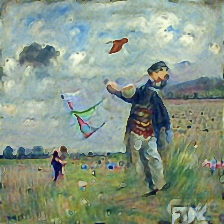} &          
      \includegraphics[width=0.18\linewidth]{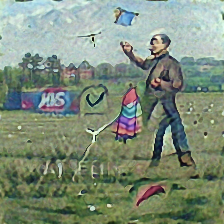} \\
            \includegraphics[width=0.18\linewidth]{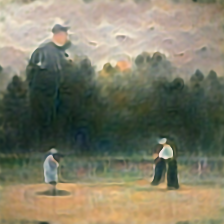} &
      \includegraphics[width=0.18\linewidth]{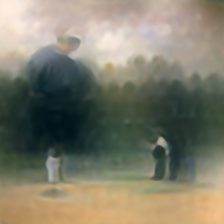} &
      \includegraphics[width=0.18\linewidth]{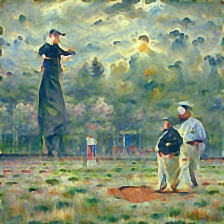} &
      \includegraphics[width=0.18\linewidth]{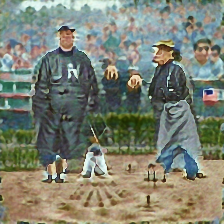} &          
      \includegraphics[width=0.18\linewidth]{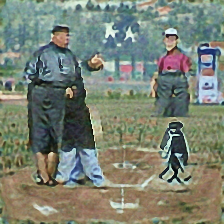} \\
            \includegraphics[width=0.18\linewidth]{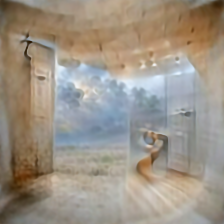} &
      \includegraphics[width=0.18\linewidth]{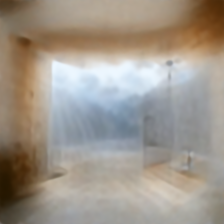} &
      \includegraphics[width=0.18\linewidth]{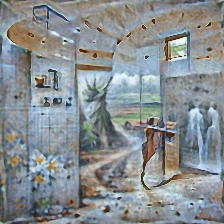} &
      \includegraphics[width=0.18\linewidth]{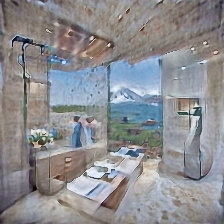} &          
      \includegraphics[width=0.18\linewidth]{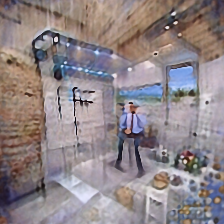} \\
            \includegraphics[width=0.18\linewidth]{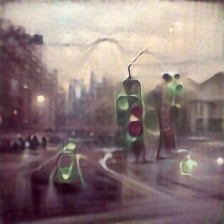} &
      \includegraphics[width=0.18\linewidth]{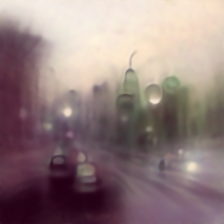} &
      \includegraphics[width=0.18\linewidth]{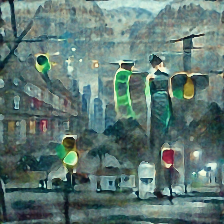} &
      \includegraphics[width=0.18\linewidth]{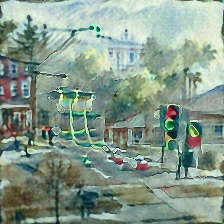} &          
      \includegraphics[width=0.18\linewidth]{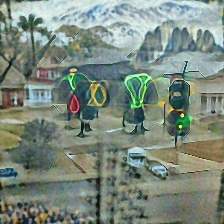} \\
            \includegraphics[width=0.18\linewidth]{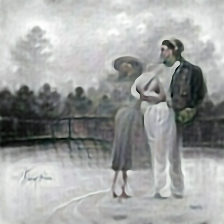} &
      \includegraphics[width=0.18\linewidth]{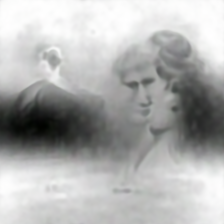} &
      \includegraphics[width=0.18\linewidth]{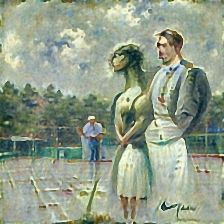} &
      \includegraphics[width=0.18\linewidth]{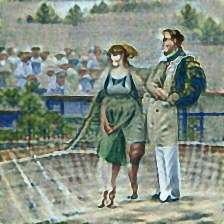} &          
      \includegraphics[width=0.18\linewidth]{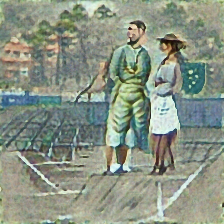} \\
            \includegraphics[width=0.18\linewidth]{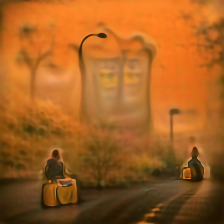} &
      \includegraphics[width=0.18\linewidth]{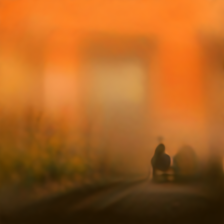} &
      \includegraphics[width=0.18\linewidth]{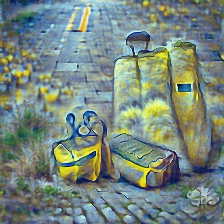} &
      \includegraphics[width=0.18\linewidth]{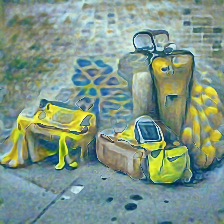} &          
      \includegraphics[width=0.18\linewidth]{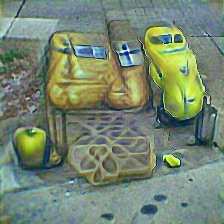} \\
    \end{tabular}
    \caption*{\small{\textbf{Right:} CLIP$^{-1}$ Ablation Rows 11-20}}
  \end{minipage}

  \caption{\small{\textbf{Ablation Study} additional samples of the ablation study shown in Figure 6 of the main paper.}}
  \label{fig:supp-qualitative}
\end{figure}

% \section{Qualitative samples of the Downstream Tasks}
% \clearpage

% \clearpage

%% file: main.bbl
\begin{thebibliography}{40}
\providecommand{\natexlab}[1]{#1}
\providecommand{\url}[1]{\texttt{#1}}
\expandafter\ifx\csname urlstyle\endcsname\relax
  \providecommand{\doi}[1]{doi: #1}\else
  \providecommand{\doi}{doi: \begingroup \urlstyle{rm}\Url}\fi

\bibitem[Betker et~al.(2023)Betker, Goh, Jing, Brooks, Wang, Li, Ouyang, Zhuang, Lee, Guo, et~al.]{betker2023improving}
James Betker, Gabriel Goh, Li Jing, Tim Brooks, Jianfeng Wang, Linjie Li, Long Ouyang, Juntang Zhuang, Joyce Lee, Yufei Guo, et~al.
\newblock Improving image generation with better captions.
\newblock \emph{Computer Science. https://cdn. openai. com/papers/dall-e-3. pdf}, 2\penalty0 (3):\penalty0 8, 2023.

\bibitem[Brock et~al.()Brock, Donahue, and Simonyan]{brocklarge}
Andrew Brock, Jeff Donahue, and Karen Simonyan.
\newblock Large scale gan training for high fidelity natural image synthesis.
\newblock In \emph{International Conference on Learning Representations}.

\bibitem[Chang et~al.(2023)Chang, Zhang, Barber, Maschinot, Lezama, Jiang, Yang, Murphy, Freeman, Rubinstein, et~al.]{chang2023muse}
Huiwen Chang, Han Zhang, Jarred Barber, Aaron Maschinot, Jose Lezama, Lu Jiang, Ming-Hsuan Yang, Kevin~Patrick Murphy, William~T Freeman, Michael Rubinstein, et~al.
\newblock Muse: Text-to-image generation via masked generative transformers.
\newblock In \emph{International Conference on Machine Learning}, pages 4055--4075. PMLR, 2023.

\bibitem[Cui et~al.(2023)Cui, Tian, Zhong, Qi, Yu, and Zhang]{cui2023decoupled}
Jiequan Cui, Zhuotao Tian, Zhisheng Zhong, Xiaojuan Qi, Bei Yu, and Hanwang Zhang.
\newblock Decoupled kullback-leibler divergence loss.
\newblock \emph{arXiv preprint arXiv:2305.13948}, 2023.

\bibitem[Esser et~al.(2024)Esser, Kulal, Blattmann, Entezari, M{\"u}ller, Saini, Levi, Lorenz, Sauer, Boesel, et~al.]{esser2024scaling}
Patrick Esser, Sumith Kulal, Andreas Blattmann, Rahim Entezari, Jonas M{\"u}ller, Harry Saini, Yam Levi, Dominik Lorenz, Axel Sauer, Frederic Boesel, et~al.
\newblock Scaling rectified flow transformers for high-resolution image synthesis.
\newblock In \emph{International Conference on Machine Learning}, pages 12606--12633. PMLR, 2024.

\bibitem[Fort and Whitaker(2025)]{fort2025direct}
Stanislav Fort and Jonathan Whitaker.
\newblock Direct ascent synthesis: Revealing hidden generative capabilities in discriminative models.
\newblock \emph{arXiv preprint arXiv:2502.07753}, 2025.

\bibitem[Frans et~al.(2022)Frans, Soros, and Witkowski]{10.5555/3600270.3600646}
Kevin Frans, L.~B. Soros, and Olaf Witkowski.
\newblock Clipdraw: exploring text-to-drawing synthesis through language-image encoders.
\newblock In \emph{Proceedings of the 36th International Conference on Neural Information Processing Systems}, Red Hook, NY, USA, 2022. Curran Associates Inc.

\bibitem[Ganz and Elad(2023)]{ganz2023clipaggeneratorfreetexttoimagegeneration}
Roy Ganz and Michael Elad.
\newblock Clipag: Towards generator-free text-to-image generation, 2023.

\bibitem[Ganz and Elad(2024)]{ganz2024texttoimagegenerationenergybasedclip}
Roy Ganz and Michael Elad.
\newblock Text-to-image generation via energy-based clip, 2024.

\bibitem[Ganz et~al.(2023)Ganz, Kawar, and Elad]{ganz2023perceptually}
Roy Ganz, Bahjat Kawar, and Michael Elad.
\newblock Do perceptually aligned gradients imply robustness?
\newblock In \emph{ICML}, 2023.

\bibitem[Hessel et~al.(2021)Hessel, Holtzman, Forbes, Bras, and Choi]{hessel2021clipscore}
Jack Hessel, Ari Holtzman, Maxwell Forbes, Ronan~Le Bras, and Yejin Choi.
\newblock Clipscore: A reference-free evaluation metric for image captioning.
\newblock \emph{arXiv preprint arXiv:2104.08718}, 2021.

\bibitem[Heusel et~al.(2017)Heusel, Ramsauer, Unterthiner, Nessler, and Hochreiter]{heusel2017gans}
Martin Heusel, Hubert Ramsauer, Thomas Unterthiner, Bernhard Nessler, and Sepp Hochreiter.
\newblock Gans trained by a two time-scale update rule converge to a local nash equilibrium.
\newblock In \emph{NeurIPS}, 2017.

\bibitem[Jagielski et~al.(2023)Jagielski, Thakkar, Tramer, Ippolito, Lee, Carlini, Wallace, Song, Thakurta, Papernot, et~al.]{jagielskimeasuring}
Matthew Jagielski, Om Thakkar, Florian Tramer, Daphne Ippolito, Katherine Lee, Nicholas Carlini, Eric Wallace, Shuang Song, Abhradeep~Guha Thakurta, Nicolas Papernot, et~al.
\newblock Measuring forgetting of memorized training examples.
\newblock In \emph{ICLR}, 2023.

\bibitem[Kazemi et~al.()Kazemi, Chegini, Geiping, Feizi, and Goldstein]{kazemi2024learn}
Hamid Kazemi, Atoosa Chegini, Jonas Geiping, Soheil Feizi, and Tom Goldstein.
\newblock What do we learn from inverting clip models?
\newblock In \emph{Neurips Safe Generative AI Workshop 2024}.

\bibitem[Kingma and Dhariwal(2018)]{kingma2018glow}
Durk~P Kingma and Prafulla Dhariwal.
\newblock Glow: Generative flow with invertible 1x1 convolutions.
\newblock \emph{Advances in neural information processing systems}, 31, 2018.

\bibitem[Liang et~al.(2022{\natexlab{a}})Liang, Zhang, Kwon, Yeung, and Zou]{NEURIPS2022_702f4db7}
Victor~Weixin Liang, Yuhui Zhang, Yongchan Kwon, Serena Yeung, and James~Y Zou.
\newblock Mind the gap: Understanding the modality gap in multi-modal contrastive representation learning.
\newblock In \emph{Advances in Neural Information Processing Systems}, pages 17612--17625. Curran Associates, Inc., 2022{\natexlab{a}}.

\bibitem[Liang et~al.(2022{\natexlab{b}})Liang, Zhang, Kwon, Yeung, and Zou]{liang2022mind}
Victor~Weixin Liang, Yuhui Zhang, Yongchan Kwon, Serena Yeung, and James~Y Zou.
\newblock Mind the gap: Understanding the modality gap in multi-modal contrastive representation learning.
\newblock \emph{Advances in Neural Information Processing Systems}, 35:\penalty0 17612--17625, 2022{\natexlab{b}}.

\bibitem[Lin et~al.(2014)Lin, Maire, Belongie, Hays, Perona, Ramanan, Doll{\'a}r, and Zitnick]{lin2014microsoft}
Tsung-Yi Lin, Michael Maire, Serge Belongie, James Hays, Pietro Perona, Deva Ramanan, Piotr Doll{\'a}r, and C~Lawrence Zitnick.
\newblock Microsoft coco: Common objects in context.
\newblock In \emph{ECCV}, 2014.

\bibitem[Liu et~al.(2024)Liu, Zhu, Zhang, Fu, Deng, Ma, Guo, and Cao]{liu2024finer}
Zhen Liu, Hao Zhu, Qi Zhang, Jingde Fu, Weibing Deng, Zhan Ma, Yanwen Guo, and Xun Cao.
\newblock Finer: Flexible spectral-bias tuning in implicit neural representation by variable-periodic activation functions.
\newblock In \emph{Proceedings of the IEEE/CVF Conference on Computer Vision and Pattern Recognition}, 2024.

\bibitem[Madry et~al.(2018)Madry, Makelov, Schmidt, Tsipras, and Vladu]{madry2017towards}
Aleksander Madry, Aleksandar Makelov, Ludwig Schmidt, Dimitris Tsipras, and Adrian Vladu.
\newblock Towards deep learning models resistant to adversarial attacks.
\newblock In \emph{ICLR}, 2018.

\bibitem[Maiorca et~al.(2023)Maiorca, Moschella, Norelli, Fumero, Locatello, and Rodol{\`a}]{maiorca2023latent}
Valentino Maiorca, Luca Moschella, Antonio Norelli, Marco Fumero, Francesco Locatello, and Emanuele Rodol{\`a}.
\newblock Latent space translation via semantic alignment.
\newblock \emph{Advances in Neural Information Processing Systems}, 36:\penalty0 55394--55414, 2023.

\bibitem[Mirza et~al.(2024)Mirza, Briglia, Beadini, and Masi]{mirza2024shedding}
Mujtaba~Hussain Mirza, Maria~Rosaria Briglia, Senad Beadini, and Iacopo Masi.
\newblock Shedding more light on robust classifiers under the lens of energy-based models.
\newblock In \emph{European Conference on Computer Vision}, pages 451--468. Springer, 2024.

\bibitem[Mistretta et~al.(2025)Mistretta, Baldrati, Agnolucci, Bertini, and Bagdanov]{mistretta2025cross}
Marco Mistretta, Alberto Baldrati, Lorenzo Agnolucci, Marco Bertini, and Andrew~D Bagdanov.
\newblock Cross the gap: Exposing the intra-modal misalignment in clip via modality inversion.
\newblock \emph{ICLR 2025}, 2025.

\bibitem[Nichol et~al.(2022)Nichol, Dhariwal, Ramesh, Shyam, Mishkin, Mcgrew, Sutskever, and Chen]{nichol2022glide}
Alexander~Quinn Nichol, Prafulla Dhariwal, Aditya Ramesh, Pranav Shyam, Pamela Mishkin, Bob Mcgrew, Ilya Sutskever, and Mark Chen.
\newblock Glide: Towards photorealistic image generation and editing with text-guided diffusion models.
\newblock In \emph{International Conference on Machine Learning}, pages 16784--16804. PMLR, 2022.

\bibitem[Radford et~al.(2015)Radford, Metz, and Chintala]{radford2015unsupervised}
Alec Radford, Luke Metz, and Soumith Chintala.
\newblock Unsupervised representation learning with deep convolutional generative adversarial networks.
\newblock \emph{arXiv preprint arXiv:1511.06434}, 2015.

\bibitem[Radford et~al.(2021)Radford, Kim, Hallacy, Ramesh, Goh, Agarwal, Sastry, Askell, Mishkin, Clark, et~al.]{radford2021learning}
Alec Radford, Jong~Wook Kim, Chris Hallacy, Aditya Ramesh, Gabriel Goh, Sandhini Agarwal, Girish Sastry, Amanda Askell, Pamela Mishkin, Jack Clark, et~al.
\newblock Learning transferable visual models from natural language supervision.
\newblock In \emph{International conference on machine learning}, pages 8748--8763. PMLR, 2021.

\bibitem[Ramesh et~al.(2022)Ramesh, Dhariwal, Nichol, Chu, and Chen]{ramesh2022hierarchical}
Aditya Ramesh, Prafulla Dhariwal, Alex Nichol, Casey Chu, and Mark Chen.
\newblock Hierarchical text-conditional image generation with clip latents.
\newblock \emph{arXiv preprint arXiv:2204.06125}, 2022.

\bibitem[Rombach et~al.(2022)Rombach, Blattmann, Lorenz, Esser, and Ommer]{rombach2022high}
Robin Rombach, Andreas Blattmann, Dominik Lorenz, Patrick Esser, and Bj{\"o}rn Ommer.
\newblock High-resolution image synthesis with latent diffusion models.
\newblock In \emph{Proceedings of the IEEE/CVF conference on computer vision and pattern recognition}, pages 10684--10695, 2022.

\bibitem[Saharia et~al.(2022)Saharia, Chan, Saxena, Li, Whang, Denton, Ghasemipour, Gontijo~Lopes, Karagol~Ayan, Salimans, et~al.]{saharia2022photorealistic}
Chitwan Saharia, William Chan, Saurabh Saxena, Lala Li, Jay Whang, Emily~L Denton, Kamyar Ghasemipour, Raphael Gontijo~Lopes, Burcu Karagol~Ayan, Tim Salimans, et~al.
\newblock Photorealistic text-to-image diffusion models with deep language understanding.
\newblock \emph{Advances in neural information processing systems}, 35:\penalty0 36479--36494, 2022.

\bibitem[Salimans et~al.(2016)Salimans, Goodfellow, Zaremba, Cheung, Radford, and Chen]{salimans2016improved}
Tim Salimans, Ian Goodfellow, Wojciech Zaremba, Vicki Cheung, Alec Radford, and Xi Chen.
\newblock Improved techniques for training gans.
\newblock \emph{Advances in neural information processing systems}, 29, 2016.

\bibitem[Sauer et~al.(2023)Sauer, Karras, Laine, Geiger, and Aila]{sauer2023stylegan}
Axel Sauer, Tero Karras, Samuli Laine, Andreas Geiger, and Timo Aila.
\newblock Stylegan-t: Unlocking the power of gans for fast large-scale text-to-image synthesis.
\newblock In \emph{International conference on machine learning}, pages 30105--30118. PMLR, 2023.

\bibitem[Schuhmann et~al.(2022)Schuhmann, Beaumont, Vencu, Gordon, Wightman, Cherti, Coombes, Katta, Mullis, Wortsman, et~al.]{schuhmann2022laion}
Christoph Schuhmann, Romain Beaumont, Richard Vencu, Cade Gordon, Ross Wightman, Mehdi Cherti, Theo Coombes, Aarush Katta, Clayton Mullis, Mitchell Wortsman, et~al.
\newblock Laion-5b: An open large-scale dataset for training next generation image-text models.
\newblock \emph{Advances in neural information processing systems}, 2022.

\bibitem[Sitzmann et~al.(2020)Sitzmann, Martel, Bergman, Lindell, and Wetzstein]{sitzmann2020implicit}
Vincent Sitzmann, Julien Martel, Alexander Bergman, David Lindell, and Gordon Wetzstein.
\newblock Implicit neural representations with periodic activation functions.
\newblock \emph{Advances in neural information processing systems}, 33:\penalty0 7462--7473, 2020.

\bibitem[Tancik et~al.(2020)Tancik, Srinivasan, Mildenhall, Fridovich-Keil, Raghavan, Singhal, Ramamoorthi, Barron, and Ng]{tancik2020fourierfeaturesletnetworks}
Matthew Tancik, Pratul~P. Srinivasan, Ben Mildenhall, Sara Fridovich-Keil, Nithin Raghavan, Utkarsh Singhal, Ravi Ramamoorthi, Jonathan~T. Barron, and Ren Ng.
\newblock Fourier features let networks learn high frequency functions in low dimensional domains, 2020.

\bibitem[Tao et~al.(2023)Tao, Bao, Tang, and Xu]{tao2023galipgenerativeadversarialclips}
Ming Tao, Bing-Kun Bao, Hao Tang, and Changsheng Xu.
\newblock Galip: Generative adversarial clips for text-to-image synthesis, 2023.

\bibitem[Wang and Mahadevan(2008)]{procrustes}
Chang Wang and Sridhar Mahadevan.
\newblock Manifold alignment using procrustes analysis.
\newblock In \emph{Proceedings of the 25th International Conference on Machine Learning}, page 1120–1127, New York, NY, USA, 2008. Association for Computing Machinery.

\bibitem[Wang et~al.(2022)Wang, Liu, He, Wu, and Yi]{wang2022clip}
Zihao Wang, Wei Liu, Qian He, Xinglong Wu, and Zili Yi.
\newblock Clip-gen: Language-free training of a text-to-image generator with clip.
\newblock \emph{arXiv preprint arXiv:2203.00386}, 2022.

\bibitem[Wu et~al.(2020)Wu, Xia, and Wang]{wu2020adversarial}
Dongxian Wu, Shu-Tao Xia, and Yisen Wang.
\newblock Adversarial weight perturbation helps robust generalization.
\newblock \emph{Advances in neural information processing systems}, 33:\penalty0 2958--2969, 2020.

\bibitem[Zhang et~al.(2019)Zhang, Yu, Jiao, Xing, Ghaoui, and Jordan]{zhang2019theoretically}
Hongyang Zhang, Yaodong Yu, Jiantao Jiao, Eric~P. Xing, Laurent~El Ghaoui, and Michael~I. Jordan.
\newblock Theoretically principled trade-off between robustness and accuracy.
\newblock In \emph{ICML}, 2019.

\bibitem[Zhang et~al.(2024)Zhang, Sui, and Yeung-Levy]{zhang2024connect}
Yuhui Zhang, Elaine Sui, and Serena Yeung-Levy.
\newblock Connect, collapse, corrupt: Learning cross-modal tasks with uni-modal data.
\newblock \emph{ICLR 2024}, 2024.

\end{thebibliography}
